\documentclass[10pt,table,twocolumn,letterpaper]{article}

\usepackage{cvpr}              

\usepackage{microtype}
\usepackage{graphicx}
\usepackage{subcaption}
\usepackage{booktabs}

\usepackage{tcolorbox}
\usepackage{hanging}
\usepackage{float}
\usepackage{enumitem}
\usepackage{multirow}
\usepackage{pifont}
\usepackage{xcolor}
\usepackage[normalem]{ulem}
\useunder{\uline}{\ul}{}
\definecolor{cvprblue}{rgb}{0.21,0.49,0.74}
\usepackage[pagebackref,breaklinks,colorlinks,allcolors=cvprblue]{hyperref}

\def\paperID{13072} 
\def\confName{CVPR}
\def\confYear{2025}

\title{Lifting the Veil on Visual Information Flow in MLLMs: Unlocking Pathways to Faster Inference}

\author{Hao Yin \quad Gunagzong Si \quad Zilei Wang\thanks{Corresponding Author} \\
University of Science and Technology of China\\
{\tt\small \{yinhnavi, guangzongsi\}@mail.ustc.edu.cn, zlwang@ustc.edu.cn}
}

\begin{document}
\maketitle

\begin{abstract}
Multimodal large language models (MLLMs) improve performance on vision-language tasks by integrating visual features from pre-trained vision encoders into large language models (LLMs). However, how MLLMs process and utilize visual information remains unclear. In this paper, a shift in the dominant flow of visual information is uncovered: (1) in shallow layers, strong interactions are observed between image tokens and instruction tokens, where most visual information is injected into instruction tokens to form cross-modal semantic representations; (2) in deeper layers, image tokens primarily interact with each other, aggregating the remaining visual information to optimize semantic representations within visual modality. Based on these insights, we propose Hierarchical Modality-Aware Pruning (HiMAP), a plug-and-play inference acceleration method that dynamically prunes image tokens at specific layers, reducing computational costs by approximately 65\% without sacrificing performance. Our findings offer a new understanding of visual information processing in MLLMs and provide a state-of-the-art solution for efficient inference. The code is available at \url{https://github.com/ustc-hyin/HiMAP}.
\end{abstract}


\section{Introduction}
Multimodal large language models (MLLMs) \cite{bai2023qwen,liu2024visual,li2023blip,zhu2023minigpt,alayrac2022flamingo} have emerged as an advanced architecture that integrates visual and textual information, demonstrating exceptional performance across various tasks. Compared to traditional multimodal models \cite{li2022blip,li2021align,li2020oscar}, MLLMs achieve superior information fusion and complex semantic understanding by utilizing large language models (LLMs) \cite{touvron2023llama,achiam2023gpt,zhao2023survey} to process visual features. However, the mechanisms of information interaction within these models remain underexplored. This study poses two critical questions: (1) To what extent do image tokens influence model predictions? (2) How is visual information processed within the model? 

With respect to the first question, we developed three metrics based on saliency scores to quantify the impact of system tokens, image tokens, and instruction tokens on prediction outcomes. Experimental results indicated that the importance of image tokens was minimal, only equivalent to 0.03\% of that of instruction tokens, despite image tokens comprising a significant portion of the model input. 

With respect to the second question, saliency analysis of the attention matrices reveals strong interactions between image tokens and instruction tokens in shallow layers, while interactions among image tokens become more significant in deeper layers. This result intuitively reveals that as the model depth increases, the dominant flow of visual information within MLLMs undergoes a shift. Based on this, we propose the following hypothesis.

\vspace{0.1cm}
\begin{tcolorbox}[title=\textit{Phased Processing of Visual Information}, colback=blue!10, colframe=blue!65!black, boxrule=0.5mm]
$\mathcal{H}_1$: \textit{In shallow layers, image tokens primarily interact with instruction tokens, injecting most visual information into instruction tokens to establish a cross-modal semantic representation for subsequent computations.}
\vspace{0.15cm} \newline
$\mathcal{H}_2$: \textit{In deeper layers, interactions among image tokens are enhanced, consolidating the residual visual information, thereby refining the semantic representation within visual modality.}
\end{tcolorbox}
\vspace{0.15cm}

\begin{figure*}[t]  
    \centering     
    \includegraphics[width=0.85\linewidth]{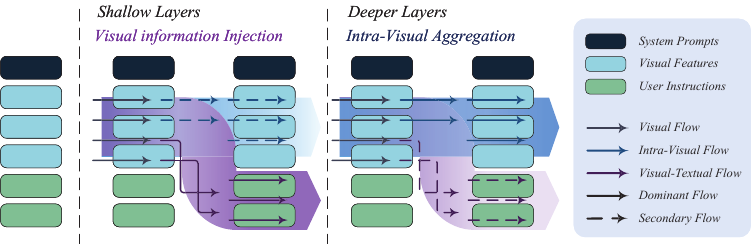} 
    \caption{Illustration of our hypothesis. In shallow layers, image tokens inject most of the visual information into instruction tokens, establishing a cross-modal semantic representation for subsequent computations. In deeper layers, image tokens aggregate the residual visual information, refining the semantic representation within the visual modality.} 
    \label{fig: phased processing of visual information} 
\end{figure*}

\cref{fig: phased processing of visual information} provides a detailed elaboration of our hypothesis. Two experiments were conducted to validate the aforementioned hypothesis. (1) By blocking the information interaction between image and instruction tokens in specific layers, we observed that perturbations in shallow layers significantly degraded model performance, confirming that image tokens inject most visual information into instruction tokens. (2) We compared the significance of \textit{visual-textual} and \textit{intra-visual information flows} at various model depths, discovering that perturbations to \textit{intra-visual information flow} in deeper layers led to more pronounced prediction deviations, thereby validating the interaction among image tokens to aggregate the residual visual information. These results support our hypothesis, indicating that MLLMs process visual information differently at varying depths.

Despite their substantial computational cost, image tokens contribute minimally to prediction results. To address this issue, we propose a method for pruning image tokens to accelerate inference. Based on insights into internal information interactions within MLLMs, we introduce Hierarchical Modality-Aware Pruning (HiMAP), a plug-and-play technique that effectively streamlines the computational process by focusing the model on the most influential image tokens. HiMAP dynamically ranks the importance of image tokens according to the dominant visual information flow at different depths and applies pruning strategies in specified layers. By reducing the computational overhead of both self-attention modules and feed-forward networks modules, HiMAP reduce FLOPs by over 65\%. Experimental results demonstrate that HiMAP can reduce inference latency by about 50\% while maintaining model performance.

In summary, our contributions are fourfold: (1) Conducting an in-depth analysis of the phenomenon where image tokens have minimal impact on prediction outcomes in MLLMs; (2) Identifying latent patterns in the interactions between visual and textual modalities within MLLMs; (3) Introducing HiMAP, a plug-and-play technique that reduces inference latency in MLLMs while maintaining performance; (4) Validating the efficacy of HiMAP across a diverse range of vision-language tasks.


\section{Inefficient Contribution of image tokens}

This section aims to highlight the limited contribution of image tokens to model predictions. \Cref{subsec: preliminaries} outlines the three token categories used as inputs in MLLMs, along with their respective processing mechanisms. \Cref{subsec: visual modality impact assessment} evaluates the impact of various modalities on prediction outcomes through metrics derived from saliency scores. The results of the quantitative analysis reveal that the visual modality contributes substantially less than other modalities.

\subsection{Preliminaries}
\label{subsec: preliminaries}

This section introduces how MLLMs process different tokens when generating output. Typically, these models follow a transformer decoder architecture \cite{vaswani2017attention}, predicting responses autoregressively \cite{brown2020language} based on a given image-question pair.

Before being fed into the transformer decoder, multimodal information (including images and text) is converted into sequence embeddings. For images, a common approach involves extracting visual features using pre-trained encoders, such as CLIP-VIT \cite{radford2021learning}. To align the dimensions of these visual features with the embedding size of LLMs and ensure semantic consistency, additional linear transformations or cross-attention modules are introduced. For text, natural language is tokenized into discrete units, and corresponding text embeddings are generated through embedding lookup. In this paper, "image tokens" and "text tokens" refer to both the discrete units of visual and textual data as well as the embeddings derived from them.

\begin{figure}[h]  
    \centering     
    \includegraphics[width=0.95 \linewidth]{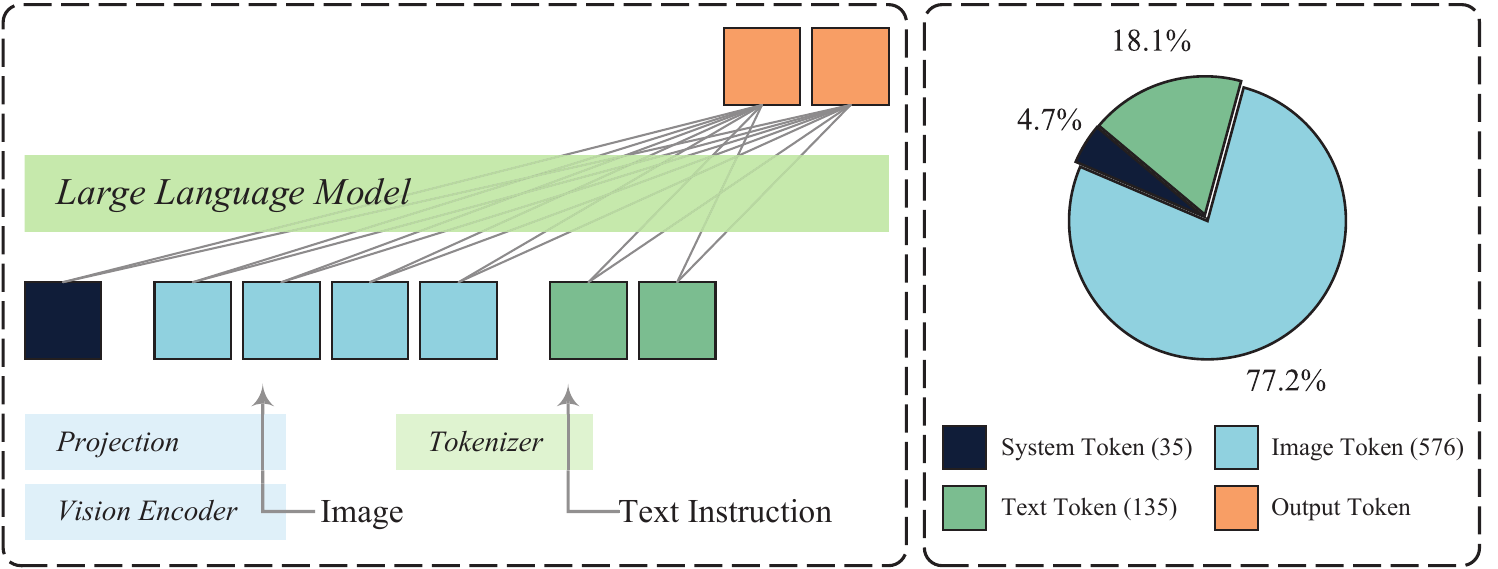} 
    \caption{Distribution of Input Sequence Tokens. Image tokens constitute 77\% of the total input tokens, nearly double the combined total of system and instruction tokens. This highlights a considerable computational overhead associated with image tokens.} 
    \label{fig: input sequence distribution} 
\end{figure}
\vspace{-0.25cm}

After preprocessing the image and text tokens into a unified embedding space, these tokens are input into the transformer decoder to generate output tokens. During this decoding process, the input tokens are categorized into three types: (1) \textbf{system prompts}, which provide general information for controlling the behavior of MLLMs; (2) \textbf{image tokens}, derived from features learned by pre-trained visual encoders; and (3) \textbf{user instructions}, which specify requests or questions related to the given images. The index sets of system, image, and instruction tokens are denoted by $\mathcal{S}$, $\mathcal{V}$, and $\mathcal{I}$, respectively. Comprehensive index set of all input tokens is represented as $\mathcal{X}$, where $\mathcal{X} = \mathcal{S} \cup \mathcal{V} \cup \mathcal{I}$.

\cref{fig: input sequence distribution} illustrates the workflow of MLLMs and the distribution of sequence lengths for three types of input tokens. The sequence length for image tokens is 576, which is nearly twice the combined length of the system and instruction tokens. This suggests that the computational load associated with image tokens in MLLMs is significantly higher.

\subsection{Visual Modality Impact Assessment}
\label{subsec: visual modality impact assessment}

This subsection quantitatively evaluates the impact of the visual modality on prediction outcomes. We employ the saliency technique \cite{simonyan2013deep}, a widely used interpretability tool, to highlight key token interactions within the attention mechanism. Following established practices, we utilize Taylor expansion~\cite{michel2019sixteen} to compute saliency scores for each element of the attention matrix:
\begin{equation}
    I_l=\left|\sum_h A_{h, l} \odot \frac{\partial \mathcal{L}(x)}{\partial A_{h, l}}\right|.
\end{equation}
Here, $A_{h, l}$ represents the attention matrix value for the $h$-th attention head in the $l$-th layer, $x$ denotes the input, and $\mathcal{L}(x)$ is the loss function of the task, e.g., the cross-entropy objective for question-answering tasks. The saliency matrix $I_l$ for the $l$-th layer is obtained by averaging across all attention heads. The significance of information flow from the $j$-th token to the $i$-th token in MLLMs is represented by $I_l(i, j)$. To illustrate the contributions of different modalities to prediction outcomes, three quantitative metrics based on $I_l$ are introduced as following.

\textbf{$\boldsymbol{S_{sys}}$, which measures the importance of information flow from system tokens to other tokens:}
\begin{equation}
    S_{sys} =\frac{1}{|\mathcal{S}|} \sum_{i \in \mathcal{X}} \sum_{j \in \mathcal{S}} I_l(i, j);
\end{equation}

\textbf{$\boldsymbol{S_{img}}$, which measures the importance of information flow from image tokens to other tokens:}
\begin{equation}
    S_{img} =\frac{1}{|\mathcal{V}|} \sum_{i \in \mathcal{X}} \sum_{j \in \mathcal{V}} I_l(i, j);
\end{equation}

\textbf{$\boldsymbol{S_{ins}}$, which measures the importance of information flow from instruction tokens to other tokens:}
\begin{equation}
    S_{ins} =\frac{1}{|\mathcal{I}|} \sum_{i \in \mathcal{X}} \sum_{j \in \mathcal{I}} I_l(i, j).
\end{equation}

These three metrics enable a systematic observation of the information flow intensity from different modalities across various layers, facilitating the evaluation of their contributions to prediction outcomes.

\begin{figure}[h]  
    \centering     
    \includegraphics[width=0.85 \linewidth]{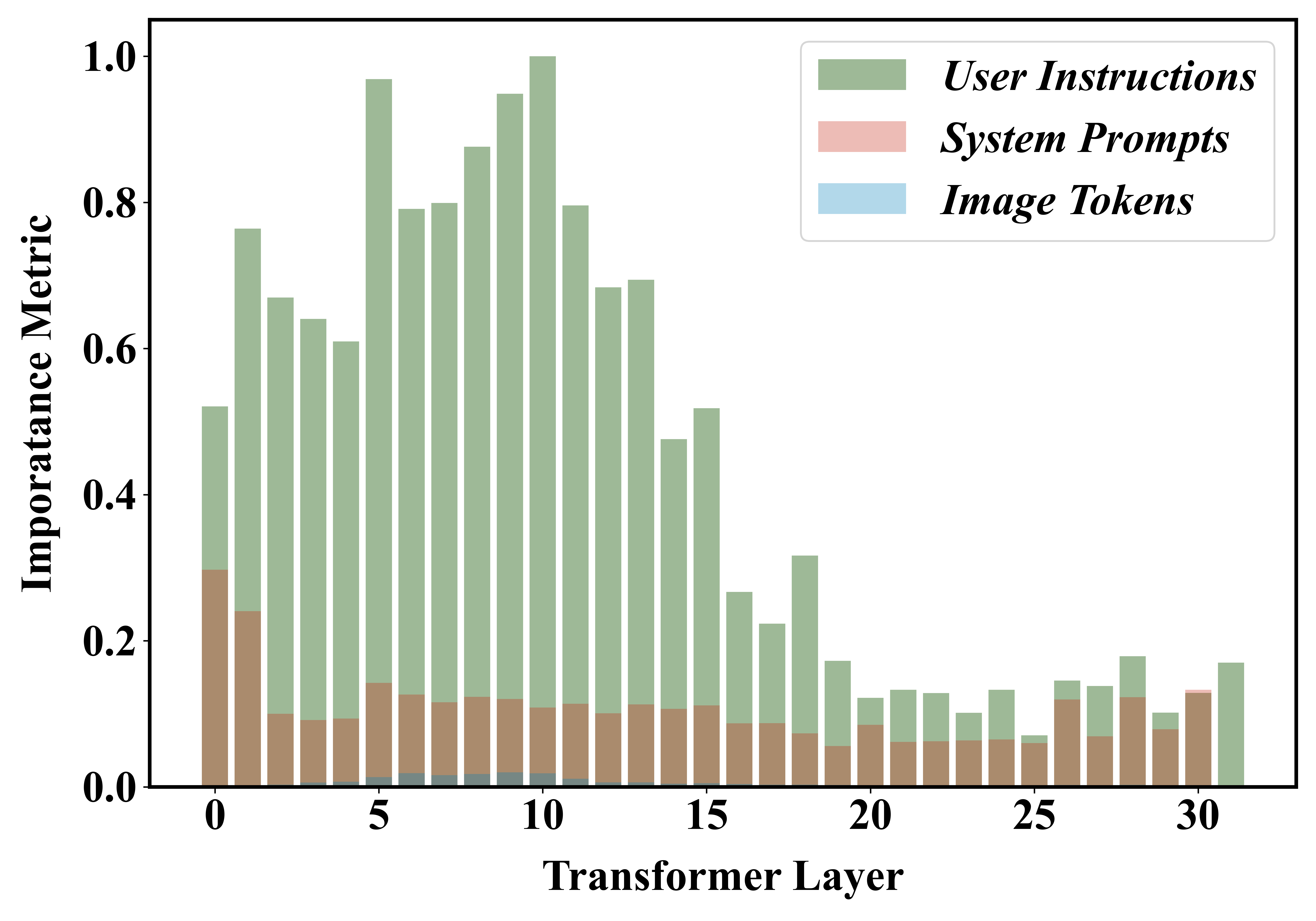} 
    \vspace{-0.25cm}
    \caption{Contributions of different modalities to prediction outcomes across layers. The contribution of visual modality is significantly lower than textual modality.} 
    \label{fig: contribution analysis} 
\end{figure}

Our experiments with the LLaVA-v1.5-7B model on A-OKVQA and Sci-VQA datasets revealed interesting insights into the impact of different modalities on prediction outcomes. As shown in \cref{fig: contribution analysis}, instruction tokens play a pivotal role in shaping predictions, exerting the most significant influence. Conversely, image tokens demonstrate a comparatively minor impact. Considering prior findings on the sparsity of attention to image tokens \cite{liu2024visualanchorsstronginformation, shang2024llavaprumergeadaptivetokenreduction,song2024low,mitra2024sparseattentionvectorsgenerative}, this limited impact may stem from the redundancy inherent in the image signals provided to the model. Additional experimental results are available in \cref{appen: modality impact assessment}.


\section{Shift in dominant flow of visual information}
This section provides a detailed analysis of how MLLMs process visual information. In \cref{subsec: hypothesis}, two importance metrics are introduced to intuitively characterize the flow of visual information within MLLMs. The quantitative results support the following hypotheses:

\begin{itemize}
    \item $\mathcal{H}_1$: \textit{In shallow layers, image tokens inject visual information into instruction tokens, facilitating cross-modal semantic representations for subsequent computations}.
    \item $\mathcal{H}_2$: \textit{In deeper layers, image tokens consolidate residual visual information, refining the semantic representation within the visual modality}.
\end{itemize}

In \cref{subsec: shallow layers} and \cref{subsec: deeper layers}, these hypotheses are validated through information flow perturbation experiments.

\subsection{Hypothesis Driven by saliency Scores}
\label{subsec: hypothesis}

This subsection seeks to uncover the underlying patterns of visual information interaction through attention mechanism in MLLMs. We continue to use $I_l(i, j)$ from Equation (1) to represent the significance of information flow from the $j$-th token to the $i$-th token. To clarify the visual information flow in MLLMs, we introduce two new quantitative metrics based on $I_l(i, j)$, with a particular focus on the information interaction involving image tokens. The metrics are defined as follows.

\textbf{$\boldsymbol{S_{vv}}$, measuring the importance of information flow among image tokens:}
\begin{equation}
    S_{vv}  = \frac{1}{|\mathcal{V}|} \sum_{ j \in \mathcal{V}} \sum_{i \in \mathcal{V}} I_l(i, j).
\end{equation}

\textbf{$\boldsymbol{S_{vt}}$, measuring the importance of information flow from image tokens to instruction tokens:}
\begin{equation}
    S_{vt}  = \frac{1}{|\mathcal{V}|} \sum_{ j \in \mathcal{I}} \sum_{i \in \mathcal{V}} I_l(i, j).
\end{equation}

$S_{vv}$ and $S_{vt}$ are utilized to analyze the mechanisms of visual information processing in MLLMs. Specifically, $S_{v t}$ quantifies the extent of information injection from image tokens to instruction tokens, whereas $S_{v v}$ measures the degree of information aggregation among image tokens. We define attention interactions among image tokens as \textit{intra-visual information flow} and those between image and instruction tokens as \textit{visual-textual information flow}.

\begin{figure}[h]
    \centering
    \includegraphics[width=0.9 \linewidth]{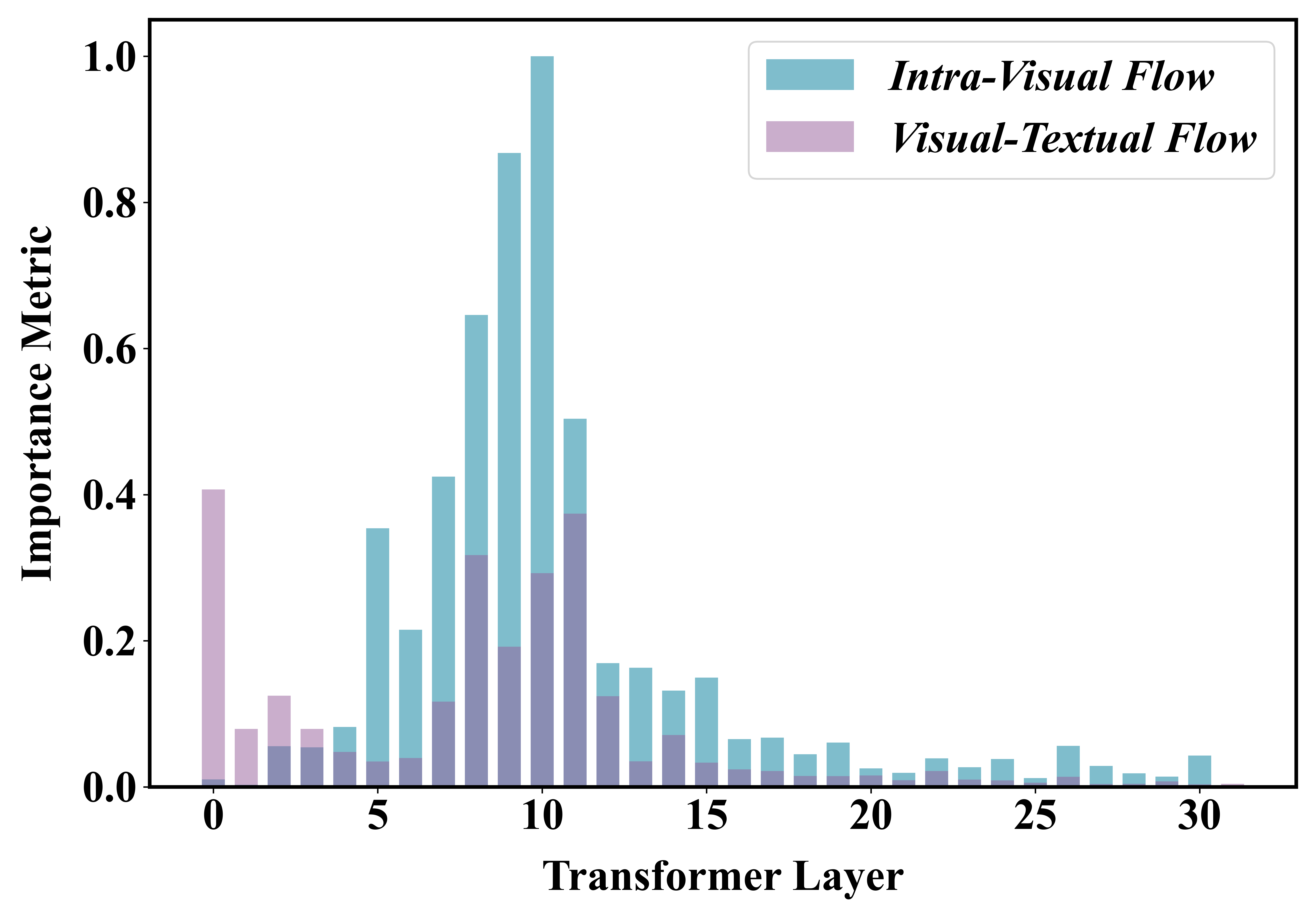} 
    \caption{Importance of \textit{intra-visual flow} and \textit{visual-textual flow} across layers. Dominant flow of visual information shifts as model depth increases.}
    \label{fig: visual flow analysis}
\end{figure}

\noindent \textbf{Results and Analysis.} We conducted experiments using the LLaVA-v1.5-7B models on the A-OKVQA and Sci-VQA datasets. As shown in \cref{fig: visual flow analysis}, the significance of two information flows changes rapidly across model depths: (1) In the shallow layers (i.e., layers 1–3), the importance of the \textit{visual-textual information flow} ($S_{vt}$) is substantially higher than that of the \textit{intra-visual information flow} ($S_{vv}$). (2) Conversely, in the deeper layers (i.e., layers 8–16), the \textit{intra-visual information flow} ($S_{vv}$) becomes predominant. Additional experimental results are detailed in \cref{appen: visual flow analysis}.

\noindent \textbf{Proposed Hypothesis.} Based on the observed shifts in dominant visual information flows, we hypothesize a phased processing mechanism for visual information in MLLMs. In the shallow layers, image tokens predominantly interact with instruction tokens, channeling the majority of visual information into these tokens to establish a cross-modal semantic representation. In the deeper layers, interactions among image tokens intensify, consolidating residual visual information to refine the semantic representation within the visual modality. This hypothesis is illustrated in \cref{fig: phased processing of visual information}.

\subsection{Shallow Layers: Visual Information Injection}
\label{subsec: shallow layers}

In this section, we validate the first part of our hypothesis. We propose that injecting visual information into instruction tokens depends on the the information flow from image tokens to instruction tokens, facilitated by the attention mechanism. By manipulating attention layers and disrupting the \textit{visual-textual information flow}, we aim to confirm the presence of this injection process and its effect on predictions.

\begin{figure}[h]
\centering
\includegraphics[width=0.95\linewidth]{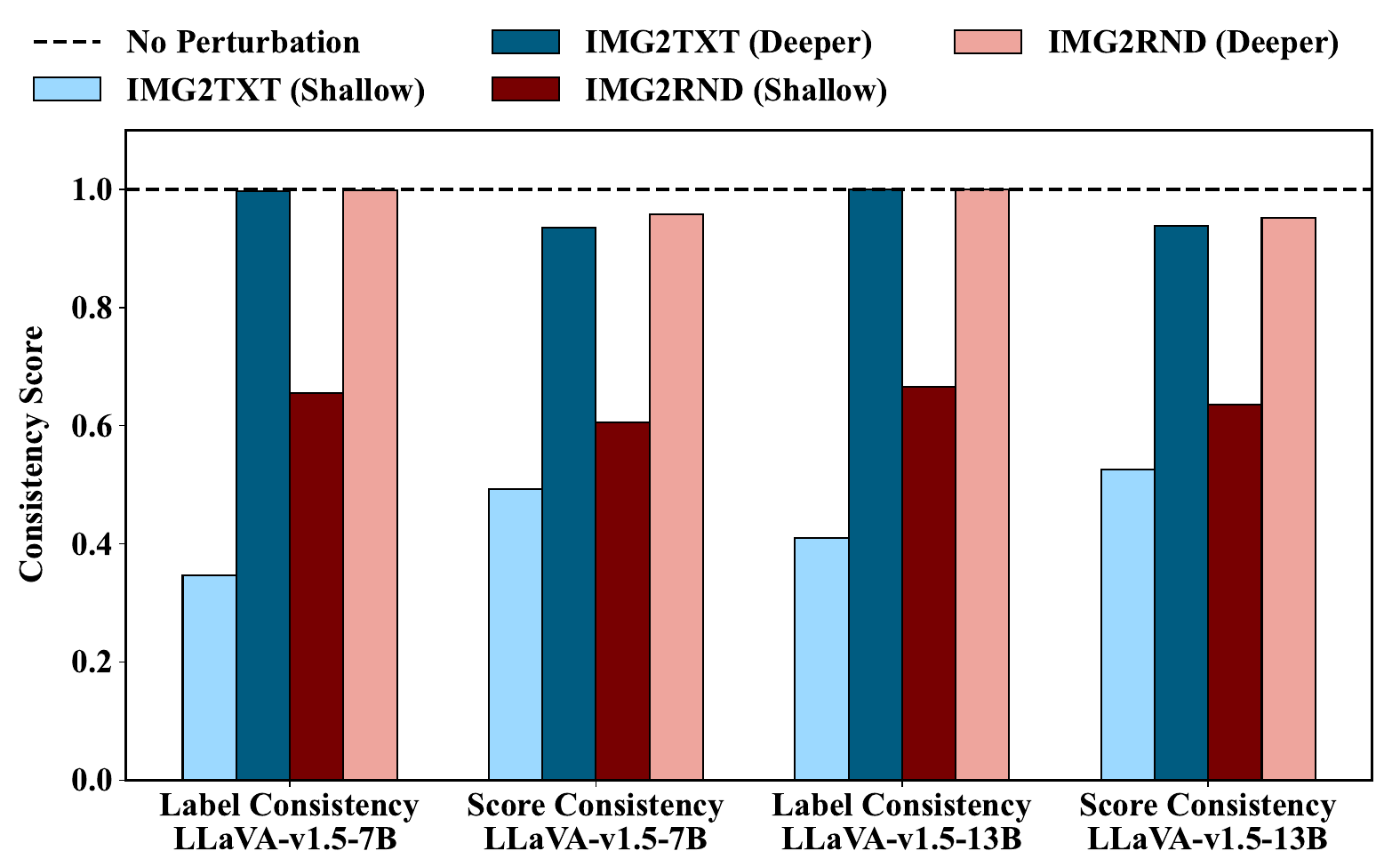} 
\caption{Disrupting \textit{visual-textual flow} versus disrupting \textit{visual-random flow} within the first or last 5 layers. Disrupting \textit{visual-textual flow} in the first 5 layers has the most substantial effect, highlighting shallow-layers information injection from image tokens to instruction tokens.}
\label{fig: disruption in shallow layers}
\end{figure}

\noindent \textbf{Implementation Details.} To disrupt the \textit{visual-textual information flow}, we block the interaction between image and instruction tokens by modifying the attention matrix $A$. Specifically, we set $A_l(i,j)$ to 0 for $i \in \mathcal{I}$ and $j \in \mathcal{V}$ in the attention matrix $A_l$ of the $l$-th layer. This modification prevents the instruction tokens from receiving information from the image tokens in the $l$-th layer.

\noindent \textbf{Evaluation Metrics.} Inspired by the loyalty metrics \cite{wang2023label}, we design the following metrics to assess the impact of disrupting \textit{visual-textual information flow}. (1) \textbf{Label Consistency}: evaluates how consistent the prediction outcomes are before and after disruption. (2) \textbf{Score Consistency}: applies the Jaccard similarity to compare the top-5 predicted tokens before and after disruption, capturing broader changes in prediction results. A lower consistency score indicates a greater impact on prediction outcomes.

\noindent \textbf{Results and Analysis.} We conducted experiments with the LLaVA-v1.5-7B models on Sci-VQA and A-OKVQA datasets. As illustrated in \cref{fig: disruption in shallow layers}, performance dropped significantly when disruptions occurred in the first five layers, but this effect diminished with increasing network depth. Conversely, disrupting the flow of information from image tokens to random tokens had only a minor impact on performance. These findings confirm that visual information is integrated into instruction tokens in the shallow layers.

\subsection{Deeper Layers: Intra-Visual Aggregation}
\label{subsec: deeper layers}

This section further validates the second part of the hypothesis, which posits that in deeper layers, enhanced interactions between image tokens lead to aggregation of residual visual information. To investigate this, we manipulated the attention layers to separately disrupt \textit{intra-visual} and \textit{visual-textual information flows}. By comparing the effects of these two disruptions on prediction outcomes, we confirm changes in the underlying visual processing mechanisms.

\begin{figure}[h]
    \centering
    \includegraphics[width=0.95\linewidth]{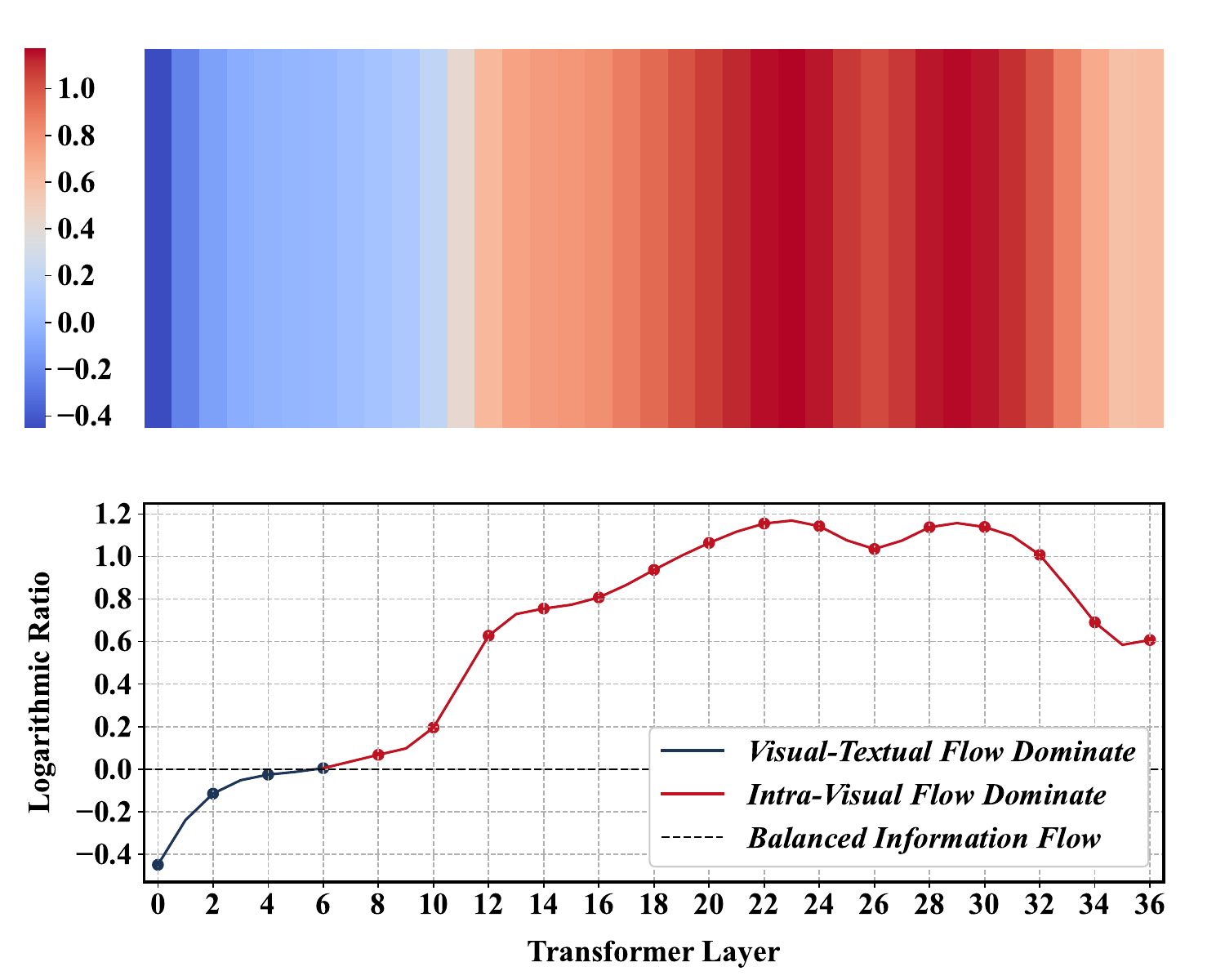} 
    \caption{The values of $D_l$ for every two layers in LLaVA-v1.5-13B. In deeper layers, $D_l > 0$, indicating that disruptions in intra-visual flow lead to greater prediction biases, thus validating the aggregation of residual visual information through interactions between image tokens.}
    \label{fig: disruption in deeper layers}
\end{figure}

\noindent \textbf{Implementation Details.} We modified the attention matrix $A$ to block interactions between image tokens, thereby disrupting \textit{intra-visual information flow}. Specifically, we set $A_l(i, j)$ to 0 for $i, j \in \mathcal{V}$ in the attention matrix $A_l$ of the $l$-th layer, thereby preventing information interactions among image tokens within that layer. The disruption in \textit{visual-textual information flow} is consistent with the procedure described in \cref{subsec: shallow layers}.

\noindent \textbf{Evaluation Metrics.} The prediction biases resulting from disruptions in \textit{visual-textual} and \textit{intra-visual information flows} are denoted as $E_{vt,l}$ and $E_{vv,l}$, respectively, where $l$ refers to the $l$-th layer. To quantify the relative impact of these two disruptions on prediction outcomes, we introduce Bias Ratio $D_l$:
\begin{equation}
    D_l = \log (E_{vv,l} / E_{vt,l}).
\end{equation}
This metric represents the logarithmic ratio of prediction biases caused by two distinct disruptions in visual information flow. When $D_l>0$, it indicates that \textit{intra-visual information flow} dominates in the $l$-th layer. Conversely, when $D_l<0$, it suggests that \textit{visual-textual information flow} prevails in the $l$-th layer. Additional experimental details are available in \cref{appen: computation of E_vv & E_vt} and \cref{appen: evaluation metrics D_l}.

\noindent \textbf{Results and Analysis} We conducted experiments using the LLaVA-v1.5-13B model on the Sci-VQA and A-OKVQA datasets. As shown in \cref{fig: disruption in deeper layers}, the results, averaged across both datasets, reveal that in the deeper layers, $D_l$ approaches 1.2, signifying the aggregation of residual visual information. Conversely, in the shallow layers, $D_l$ drops to -0.5, indicating the integration of visual information from image tokens into instruction tokens.

\subsection{Hypothesis Discussion}
In \cref{subsec: shallow layers}, we establish that image tokens predominantly transfer visual information to instruction tokens in the shallow layers. In contrast, as detailed in \cref{subsec: deeper layers}, image tokens aggregate residual visual information in the deeper layers. Furthermore, \cref{subsec: visual modality impact assessment} highlights the inefficiency of image tokens in influencing prediction outcomes. Based on these findings, \textit{we propose that not all image tokens are necessary at every layer}. Specifically, once the majority of visual information has been injected into instruction tokens in the shallow layers, many image tokens lose their significance. This explains why the \textit{visual-textual information flow} has minimal influence on prediction outcomes in the subsequent layers. Similarly, in the deeper layers, after aggregating the residual visual information, most remaining image tokens become redundant.


\section{Hierarchical Modality-Aware Pruning}
Given the relatively minor contribution of image tokens to prediction outcomes, coupled with their substantial computational cost in MLLMs, we propose a dynamic pruning method for image tokens. This approach effectively reduces computational overhead during inference without compromising model performance.

\subsection{Hierarchical image token Pruning}

Leveraging the phased processing mechanism of visual information in MLLMs, we propose Hierarchical Modality-Aware Pruning (HiMAP), a technique designed to accelerate inference by dynamically pruning image tokens at varying network depths.

\begin{figure}[h]  
    \centering     
    \includegraphics[width=0.98 \linewidth]{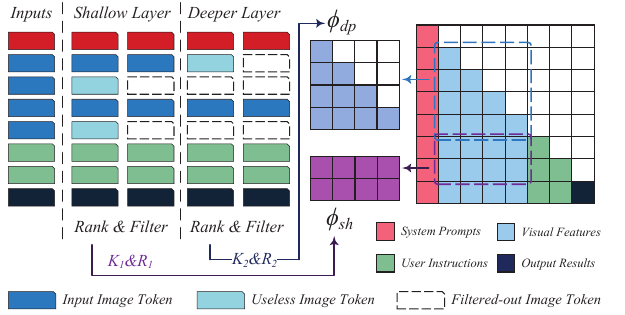} 
    \caption{Illustration of Hierarchical Modality-Aware Pruning (HiMAP). In shallow layers, HiMAP ranks image tokens at the $K_1$-th layer based on the importance criterion $\phi_{sh}$, removing the image tokens in the bottom $R_1\%$. In deeper layers, HiMAP ranks the remaining image tokens at the $K_2$ layer according to the importance criterion $\phi_{dp}$, filtering out those in the bottom $R_2\%$.} 
    \label{fig: himap} 
\end{figure}

\cref{fig: himap} illustrates the overall framework of HiMAP, which includes two core components: shallow-layer pruning module and deeper-layer pruning module. Each module features an importance ranking function $f_\phi$ and two parameters: the filtering layer $K$ and the filtering ratio $R \%$. At the $K$-th layer of MLLMs, the ranking function $f_\phi$ accepts a set of image tokens as input and ranks them according to a predefined importance criterion $\phi$. After ranking, image tokens deemed to have the lowest importance in the bottom $R \%$ are pruned in subsequent layers, thus optimizing the utilization of computational resources.

In shallow layers, image tokens primarily interact with instruction tokens, injecting most visual information into instruction tokens. Consequently, we define the importance criterion $\phi_{sh}$ in shallow-layer pruning module as the sum of attention scores from the given image token $v$ to all instruction tokens, represented as:
\begin{equation}
    \phi_{sh} (v) = \sum_{i \in \mathcal{I}} A_{K_1}(i, v).
\end{equation}
Here, $K_1$ denotes the filtering layer in shallow-layer pruning module. This criterion quantifies the influence of image tokens on instruction tokens, thereby guiding the pruning of image tokens in shallow layers.

In deeper layers, interactions among image tokens are enhanced, consolidating the residual visual information. Thus, we define the importance criterion $\phi_{dp}$ in deeper-layer pruning module as the sum of attention scores from all other image tokens to the given image token $v$, expressed as:
\begin{equation}
    \phi_{dp} (v) = \sum_{i \in \mathcal{V}} A_{K_2}(v, i)
\end{equation}
Here, $K_2$ denotes the filtering layer in deeper-layer pruning module. This criterion evaluates the information interaction between image tokens, directing the pruning of image tokens in deeper layers.

\subsection{Computation Cost Estimation}

We estimate the computational cost of the multi-head attention (MHA) and feed-forward neural network (FFN) modules in terms of Floating Point Operations Per Second (FLOPs). For a transformer layer, given that the input contains $n$ image tokens, the hidden layer dimension is $d$, and the intermediate layer dimension of the FFN is $m$, the FLOPs for this layer can be represented as $\Omega(n)=4 n d^2+2 n^2 d+2 n d m$.

For the entire model, assuming there are $L$ layers in total, the shallow-layer pruning module reduces the number of image tokens from $n$ to $n_1 = (1 - R_{1}\%) \cdot n$ at the $K_1$-th layer. The deeper-layer pruning module further reduces the number of image tokens to $n_2 = (1 - R_{2}\%) \cdot n_1$. Thus, the theoretical FLOPs reduction rate $\eta$ related to image tokens can be calculated using the following formula:

\begin{equation}
\begin{split}
    \eta = 1 - \frac{K_1 \cdot \Omega(n) + (K_2-K_1) \cdot  \Omega(n_1) }{L \cdot  \Omega(n)} \\
    - \frac{(L-K_2) \cdot  \Omega(n_2)}{L \cdot  \Omega(n)}
\end{split}
\end{equation}

\begin{table*}[t]
\small
\centering
\label{tab: short QA}
\begin{tabular}{@{}c|c|cc|cccc|cc@{}}
\toprule[1pt]
\toprule
\textbf{Model}                & \textbf{Method}               & \textbf{TFLOPs}                                              & \textbf{FLOPs Ratio}                                         & \textbf{VQAv2}                                               & \textbf{T-VQA}                                               & \textbf{POPE}                         & \textbf{MME}                                                   & \textbf{S-VQA}                                               & \textbf{A-OKVQA}                                             \\ \midrule
                              & Baseline                      & 2.98                                                         & 100\%                                                        & \textbf{78.3}                                                & 58.2                                                         & {\color[HTML]{CB0000} \textbf{86.4}}  & \textbf{1749.9}                                                & 67.9                                                         & 76.6                                                         \\
                              & FastV                         & 1.56                                                         & 54\%                                                         & 78.1                                                         & \textbf{58.3}                                                & 84.9                                  & 1742.6                                                         & \textbf{68.1}                                                & \textbf{77}                                                  \\
\multirow{-3}{*}{LLaVA-7B}    & \cellcolor[HTML]{ECF4FF}HiMAP & \cellcolor[HTML]{ECF4FF}{\color[HTML]{036400} \textbf{0.73}} & \cellcolor[HTML]{ECF4FF}{\color[HTML]{036400} \textbf{24\%}} & \cellcolor[HTML]{ECF4FF}{\color[HTML]{CB0000} \textbf{78.6}} & \cellcolor[HTML]{ECF4FF}{\color[HTML]{CB0000} \textbf{58.4}} & \cellcolor[HTML]{ECF4FF}\textbf{86.2} & \cellcolor[HTML]{ECF4FF}{\color[HTML]{CB0000} \textbf{1785.1}} & \cellcolor[HTML]{ECF4FF}{\color[HTML]{CB0000} \textbf{68.3}} & \cellcolor[HTML]{ECF4FF}{\color[HTML]{CB0000} \textbf{77.2}} \\ \midrule
                              & Baseline                      & 5.81                                                         & 100\%                                                        & 79.8                                                         & 61.4                                                         & {\color[HTML]{CB0000} \textbf{87.2}}  & 1794.4                                                         & \textbf{71.6}                                                & {\color[HTML]{CB0000} \textbf{82}}                           \\
                              & FastV                         & 3.09                                                         & 53\%                                                         & \textbf{79.9}                                                & \textbf{61.4}                                                & 84.8                                  & \textbf{1796.3}                                                & 71.3                                                         & 81.3                                                         \\
\multirow{-3}{*}{LLaVA-13B}   & \cellcolor[HTML]{ECF4FF}HiMAP & \cellcolor[HTML]{ECF4FF}{\color[HTML]{036400} \textbf{1.36}} & \cellcolor[HTML]{ECF4FF}{\color[HTML]{036400} \textbf{23\%}} & \cellcolor[HTML]{ECF4FF}{\color[HTML]{CB0000} \textbf{80.2}} & \cellcolor[HTML]{ECF4FF}{\color[HTML]{CB0000} \textbf{61.7}} & \cellcolor[HTML]{ECF4FF}\textbf{86.5} & \cellcolor[HTML]{ECF4FF}{\color[HTML]{CB0000} \textbf{1809.4}} & \cellcolor[HTML]{ECF4FF}{\color[HTML]{CB0000} \textbf{72.1}} & \cellcolor[HTML]{ECF4FF}{\color[HTML]{333333} \textbf{81.4}} \\ \midrule
                              & Baseline                      & 3.6                                                          & 100\%                                                        & 78.4                                                         & \textbf{60.8}                                                & {\color[HTML]{CB0000} \textbf{84.5}}  & \textbf{1782.6}                                                & 68                                                           & 75.7                                                         \\
                              & FastV                         & 1.9                                                          & 53\%                                                         & \textbf{78.5}                                                & 58.3                                                         & 82.7                                  & 1767.2                                                         & \textbf{68.2}                                                & \textbf{75.3}                                                \\
\multirow{-3}{*}{QwenVL-7B}   & \cellcolor[HTML]{ECF4FF}HiMAP & \cellcolor[HTML]{ECF4FF}{\color[HTML]{036400} \textbf{0.89}} & \cellcolor[HTML]{ECF4FF}{\color[HTML]{036400} \textbf{25\%}} & \cellcolor[HTML]{ECF4FF}{\color[HTML]{CB0000} \textbf{78.8}} & \cellcolor[HTML]{ECF4FF}{\color[HTML]{CB0000} \textbf{61.3}} & \cellcolor[HTML]{ECF4FF}\textbf{83.7} & \cellcolor[HTML]{ECF4FF}{\color[HTML]{CB0000} \textbf{1798.3}} & \cellcolor[HTML]{ECF4FF}{\color[HTML]{CB0000} \textbf{68.5}} & \cellcolor[HTML]{ECF4FF}{\color[HTML]{CB0000} \textbf{75.9}} \\ \midrule
                              & Baseline                      & 2.71                                                         & 100\%                                                        & \textbf{79.4}                                                & \textbf{57.1}                                                & {\color[HTML]{CB0000} \textbf{86.9}}  & \textbf{1812.2}                                                & \textbf{70.9}                                                & \textbf{79.6}                                                \\
                              & FastV                         & 1.39                                                         & 52\%                                                         & 79                                                           & 56.8                                                         & 85.2                                  & 1802.1                                                         & 70.5                                                         & 79.1                                                         \\
\multirow{-3}{*}{InternVL-7B} & \cellcolor[HTML]{ECF4FF}HiMAP & \cellcolor[HTML]{ECF4FF}{\color[HTML]{036400} \textbf{0.56}} & \cellcolor[HTML]{ECF4FF}{\color[HTML]{036400} \textbf{20\%}} & \cellcolor[HTML]{ECF4FF}{\color[HTML]{CB0000} \textbf{79.6}} & \cellcolor[HTML]{ECF4FF}{\color[HTML]{CB0000} \textbf{57.1}} & \cellcolor[HTML]{ECF4FF}\textbf{86.5} & \cellcolor[HTML]{ECF4FF}{\color[HTML]{CB0000} \textbf{1821.3}} & \cellcolor[HTML]{ECF4FF}{\color[HTML]{CB0000} \textbf{71.7}} & \cellcolor[HTML]{ECF4FF}{\color[HTML]{CB0000} \textbf{80.1}} \\ \bottomrule
\bottomrule[1pt]
\end{tabular}
\vspace{0.2cm}
\caption{\textbf{Performance of HiMAP on Short-answer QA and Multiple-choice QA Tasks}. The parameters are set as $K_1 = 2, R_1 = 50\%$, $K_2 = 8, R_2 = 75\%$. The evaluation metric used is \textbf{Accuracy}. The {\color[HTML]{CB0000} \textbf{highest}} score for each configuration is highlighted in {\color[HTML]{CB0000} \textbf{red}}, while the {\color[HTML]{036400} \textbf{lowest}} computational cost is marked in {\color[HTML]{036400} \textbf{green}}.}
\end{table*}

\section{Experiments}

In this section, we showcase HiMAP's performance across various visual language benchmarks. \cref{subsec: evaluation tasks & experimental settings} provides an overview of the evaluation tasks and parameter configurations. \cref{subsec: experimental results} highlights HiMAP's exceptional results in terms of both inference accuracy and speed. Lastly, \cref{subsec: ablation study} presents ablation studies that validate the effectiveness of HiMAP's individual components.

\subsection{Experimental Setup}
\label{subsec: evaluation tasks & experimental settings}

The performance of the HiMAP method is rigorously examined across four essential tasks: short-answer QA, multiple-choice QA, image captioning, and natural QA. These tasks encompass structured question-answering and open-ended generation, offering a comprehensive evaluation of MLLMs utilizing HiMAP method.

\begin{itemize}
    \item \textbf{Short-answer QA:} This task utilizes datasets such as VQAv2 \cite{goyal2017makingvvqamatter}, TextVQA \cite{singh2019vqamodelsread}, MME \cite{fu2024mmecomprehensiveevaluationbenchmark}, and POPE \cite{li2023evaluating} to assess the model’s ability to generate concise answers from multimodal information.
    \item \textbf{Multiple-choice QA:} Using ScienceQA \cite{lu2022learn} and A-OKVQA \cite{schwenk2022okvqa} datasets, this task tests reasoning and decision-making in multiple-choice scenarios.
    \item \textbf{Image Captioning:} Nocaps \cite{agrawal2019nocaps} and Flickr30k \cite{plummer2015flickr30k, plummer2016flickr30kentitiescollectingregiontophrase} datasets are employed to measure the model’s ability to generate natural language descriptions of images. Performance is evaluated using the CIDEr score \cite{vedantam2015cider}.
    \item \textbf{Natural QA:} This task involves experiments with the LLaVA-Bench \cite{liu2024visual} and MM-Vet \cite{yu2023mm} datasets. GPT-4 \cite{achiam2023gpt} is used to assess the correctness of the generated answers, evaluating the model’s capability to handle open-ended question-answering in complex multimodal scenarios.
\end{itemize}

\noindent Experiments were done with mainstream MLLMs, including LLaVA-v1.5-7B \cite{liu2024improvedbaselinesvisualinstruction}, LLaVA-v1.5-13B, QwenVL-Chat-7B \cite{bai2023qwen}, and InternVL-v1.0-7B \cite{chen2024internvl, chen2024far}. For structured question-answering tasks such as Short-answer QA and Multiple-choice QA, HiMAP adopts an aggressive parameter configuration ($K_1=2, R_1=50\%, K_2=8, R_2=75\%$) to effectively prune irrelevant image tokens, significantly enhancing inference speed. Conversely, for open-ended generation tasks like Image Captioning and Natural QA, HiMAP applies a more conservative parameter configuration ($K_1=2, R_1=50\%, K_2=15, R_2=75\%$) to ensure that the model captures comprehensive image information, leading to high-quality responses.

We primarily compared our method to FastV \cite{chen2024image}, which calculates the average attention score received by a token from all other tokens as its importance criterion, denoted as $\phi_{attn}$. For a fair comparison, we adopted the optimal settings recommended for FastV: $K=2$ and $R=50\%$ for LLaVA-v1.5-7B, and $K=3$ and $R=50\%$ for LLaVA-v1.5-13B.

\subsection{Results and Analysis}
\label{subsec: experimental results}

As shown in \cref{tab: short QA}, HiMAP achieves outstanding accuracy in both Short-answer QA and Multiple-choice QA tasks. In \textbf{Short-answer QA} task, HiMAP surpasses baseline accuracy on VQAv2, TextVQA, and MME datasets, demonstrating its superior precision and stability in pruning image tokens. Even on the POPE benchmark, where performance declines for both HiMAP and FastV, HiMAP exhibits a notably smaller impact. 

Similarly, in \textbf{Multiple-choice QA} task, HiMAP outperforms the baseline. By contrast, applying FastV to the InternVL-v1.0-7B and LLaVA-v1.5-13B models results in significant performance degradation, further underscoring HiMAP’s advantages. Additional experimental results for \textbf{Short-answer QA} tasks can be found in \cref{appen: ChartQA Doc-QA} and \cref{appen: mme}.

\begin{table}[h]
\small
\centering
\label{tab: image caption}
\begin{tabular}{@{}c|cc|cc@{}}
\toprule[1pt]
\toprule
\textbf{Model}              & \textbf{Method}               & \textbf{Ratio}                                               & \textbf{Nocaps}                                              & \textbf{Flickr30k}                                           \\ \midrule
                            & Baseline                      & 100\%                                                        & \textbf{78.8}                                                & \textbf{50.9}                                                \\
                            & FastV                         & 54\%                                                         & 78.6                                                         & 50.6                                                         \\
\multirow{-3}{*}{LLaVA-7B}  & \cellcolor[HTML]{ECF4FF}HiMAP & \cellcolor[HTML]{ECF4FF}{\color[HTML]{036400} \textbf{34\%}} & \cellcolor[HTML]{ECF4FF}{\color[HTML]{CB0000} \textbf{78.7}} & \cellcolor[HTML]{ECF4FF}{\color[HTML]{CB0000} \textbf{51.3}} \\ \midrule
                            & Baseline                      & 100\%                                                        & \textbf{82.8}                                                & \textbf{53.6}                                                \\
                            & FastV                         & 53\%                                                         & 81.9                                                         & 53.1                                                         \\
\multirow{-3}{*}{LLaVA-13B} & \cellcolor[HTML]{ECF4FF}HiMAP & \cellcolor[HTML]{ECF4FF}{\color[HTML]{036400} \textbf{33\%}} & \cellcolor[HTML]{ECF4FF}{\color[HTML]{CB0000} \textbf{83.7}} & \cellcolor[HTML]{ECF4FF}{\color[HTML]{CB0000} \textbf{53.8}} \\ \bottomrule
\bottomrule[1pt]
\end{tabular}
\vspace{0.15cm}
\caption{\textbf{Performance of HiMAP on Image Captioning task}. The parameters for HiMAP are set as $K_1=2, R_1=50 \%, K_2=15$, and $R_2=75 \%$. The evaluation metric used is the \textbf{CIDEr} score. The {\color[HTML]{CB0000} \textbf{highest}} score for each configuration is highlighted in {\color[HTML]{CB0000} \textbf{red}}, while the {\color[HTML]{036400} \textbf{lowest}} computational cost is marked in {\color[HTML]{036400} \textbf{green}}.}
\end{table}

As shown in \cref{tab: image caption}, HiMAP performs comparably to the baseline in \textbf{Image Captioning} task, whereas FastV results in a noticeable decline in model performance. This clearly demonstrates that HiMAP is better at preserving the model's ability to generate long-text descriptions compared to FastV.

\begin{figure*}[t]
    \centering
    \begin{subfigure}[t]{0.24\linewidth}
        \centering
        \includegraphics[width=\linewidth]{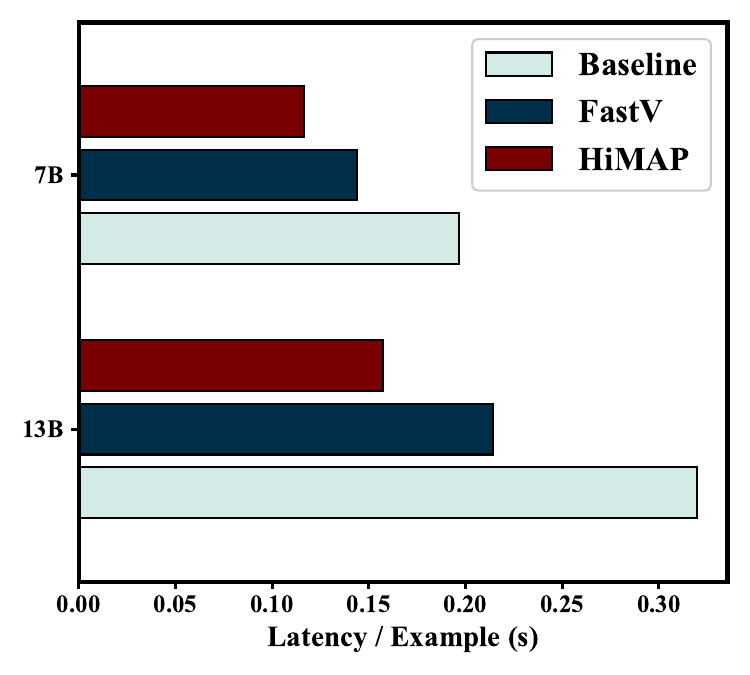}
        \caption{ScienceQA Dataset}
        \label{fig: scivqa}
    \end{subfigure}
    \hfill
    \begin{subfigure}[t]{0.24\linewidth}
        \centering
        \includegraphics[width=\linewidth]{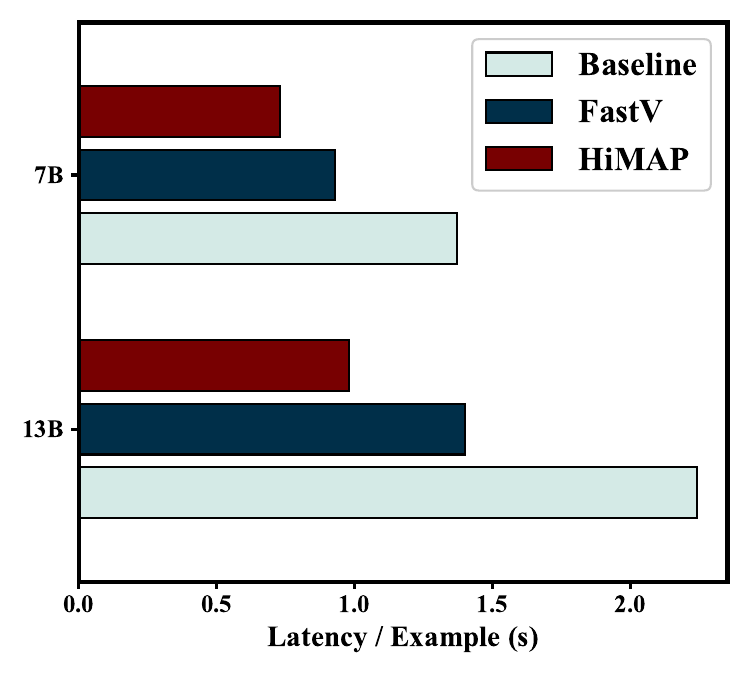}
        \caption{A-OKVQA Dataset}
        \label{fig: aokvqa}
    \end{subfigure}
    \hfill
    \begin{subfigure}[t]{0.24\linewidth}
        \centering
        \includegraphics[width=\linewidth]{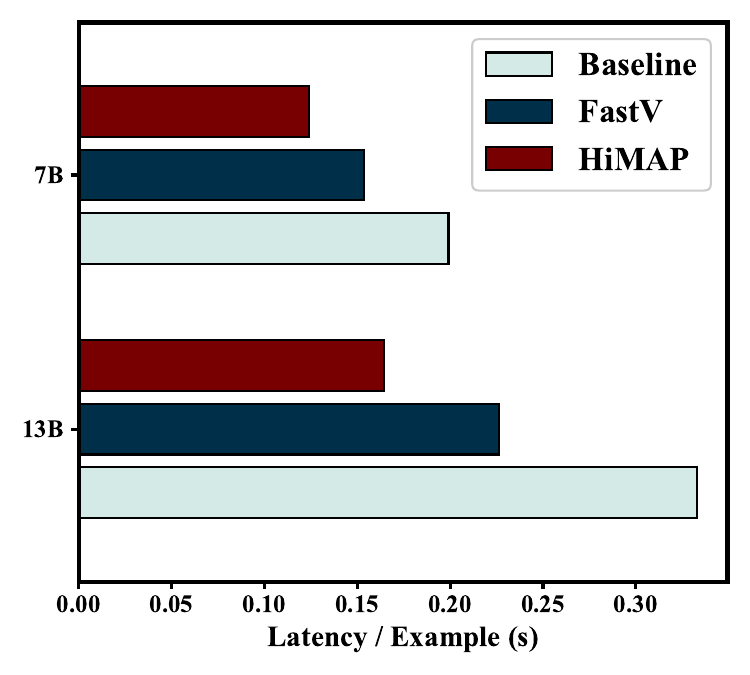}
        \caption{Nocaps Dataset}
        \label{fig: nocaps}
    \end{subfigure}
    \hfill
    \begin{subfigure}[t]{0.24\linewidth}
        \centering
        \includegraphics[width=\linewidth]{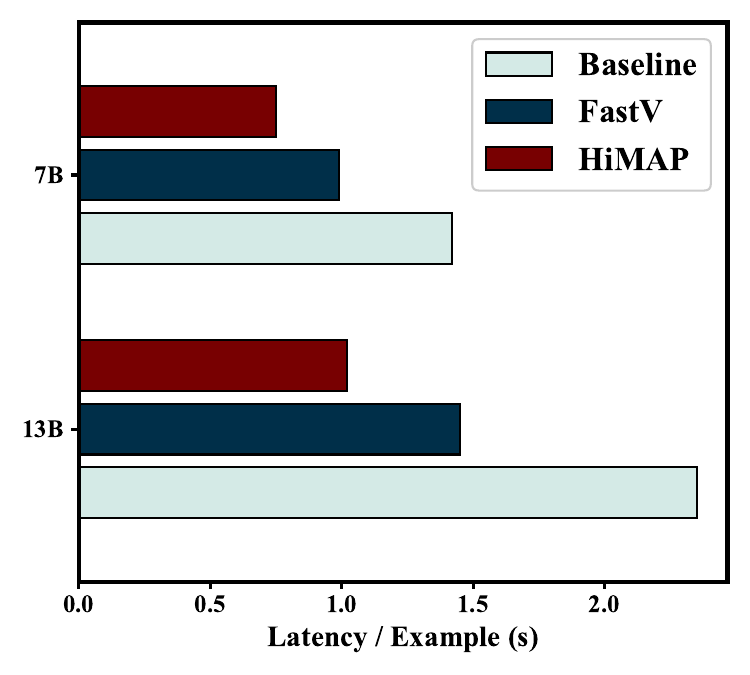}
        \caption{Flickr30k Dataset}
        \label{fig: flickr30k}
    \end{subfigure}
    \caption{\textbf{Comparison of real-world inference speeds between HiMAP and FastV.} The experiment was conducted using LLaVA-v1.5 model family on a server equipped with a single 80GB A800 GPU.}
    \label{fig: inference speed}
\end{figure*}

\cref{tab: Natural QA} presents the experimental results of HiMAP and FastV on \textbf{Natural QA} task. Based on GPT-4's evaluation, HiMAP's responses significantly surpass those of FastV, even exceeding the baseline. This highlights HiMAP's ability to maintain robust open-domain QA performance in complex scenarios. For case studies on LLaVA-Bench, please see \cref{appen: case study}.

\begin{table}[h]
\small
\centering
\begin{tabular}{@{}c|cc|cc@{}}
\toprule[1pt]
\toprule
\textbf{Model}              & \textbf{Method} & \textbf{Ratio}                                               & \textbf{LV-BM}                                               & \textbf{MM-Vet}                                              \\ \midrule
                            & Baseline        & 100\%                                                        & \textbf{65.7}                                                & \textbf{33.4}                                                \\
                            & FastV           & 54\%                                                         & 62.4                                                         & 31.2                                                         \\
\multirow{-3}{*}{LLaVA-7B}  &\cellcolor[HTML]{ECF4FF} HiMAP           & \cellcolor[HTML]{ECF4FF}{\color[HTML]{036400} \textbf{34\%}} & \cellcolor[HTML]{ECF4FF}{\color[HTML]{CB0000} \textbf{66.5}} & \cellcolor[HTML]{ECF4FF}{\color[HTML]{CB0000} \textbf{33.7}} \\ \midrule
                            & Baseline        & 100\%                                                        & \textbf{73.5}                                                & {\color[HTML]{333333} \textbf{37.4}}                         \\
                            & FastV           & 53\%                                                         & 71.7                                                         & 35.5                                                         \\
\multirow{-3}{*}{LLaVA-13B} &\cellcolor[HTML]{ECF4FF} HiMAP           & \cellcolor[HTML]{ECF4FF}{\color[HTML]{036400} \textbf{33\%}} & \cellcolor[HTML]{ECF4FF}{\color[HTML]{CB0000} \textbf{74.5}} & \cellcolor[HTML]{ECF4FF}{\color[HTML]{CB0000} \textbf{37.4}} \\ \bottomrule
\bottomrule[1pt]
\end{tabular}
\caption{\textbf{Performance of HiMAP on Natural QA task}. The parameters for HiMAP are set as $K_1=2, R_1=50\%, K_2=15, R_2=75\%$. LV-BM refers to LLaVA-Bench dataset. The {\color[HTML]{CB0000} \textbf{highest}} score for each configuration is highlighted in {\color[HTML]{CB0000} \textbf{red}}, while the {\color[HTML]{036400} \textbf{lowest}} computational cost is marked in {\color[HTML]{036400} \textbf{green}}.}
\label{tab: Natural QA}
\end{table}

As shown in \cref{tab: short QA}, for \textbf{Short-answer QA} and \textbf{Multiple-choice QA} tasks, HiMAP adopts aggressive parameter settings, specifically $K_1=2, R_1=50 \%, K_2=8, R_2=75 \%$ . These settings achieve approximately a $75 \%$ reduction in FLOPs without sacrificing model performance, significantly decreasing computational overhead during inference.

For \textbf{Image Captioning} and \textbf{Natural QA} tasks, as illustrated in \cref{tab: image caption,tab: Natural QA}, HiMAP employs more conservative parameter configurations, namely $K_1=2, R_1=50 \%, K_2=15, R_2=$ $75 \%$. Despite the more restrained settings, HiMAP still achieves a roughly 70\% reduction in computational costs, significantly outperforming FastV's 45\% reduction rate.

\cref{fig: inference speed} compares the real-world inference speed of HiMAP and FastV. The experiment was conducted on a server equipped with a single 80GB A800 GPU. HiMAP achieved a 60\% reduction in inference time, significantly surpassing FastV’s 35\% improvement. Details on GPU memory usage and more comprehensive evaluations of inference speed can be found in \cref{appen: inference speed}.

\subsection{Ablation Study}
\label{subsec: ablation study}

We conducted ablation experiments using the LLaVA-v1.5-7B model, as summarized in \cref{tab: ablation study}. Pruning visual tokens in shallow layers based on intra-visual information flow caused a significant performance drop, confirming their role in integrating visual information with instruction tokens for cross-modal semantic representations. Similarly, pruning tokens in deeper layers, guided by visual-textual information flow, also degraded performance, supporting their function in refining semantic representations within the visual modality. For additional ablation experiments, please refer to \cref{appen: ablation study}.

\begin{table}[h]
\small
\centering
\begin{tabular}{@{}cc|c|cc@{}}
\toprule[1pt]
\toprule
\textbf{SHL-PM}   & \textbf{DPL-PM}   & \textbf{Ratio}                        & \textbf{ScienceQA}                      & \textbf{A-OKVQA}                     \\ \midrule
\ding{55}        & \ding{55}        & {\color[HTML]{CB0000} \textbf{100\%}} & 67.9                                 & 76.6                                 \\
\textit{img2txt} & \ding{55}        & {\color[HTML]{FE0000} \textbf{54\%}}  & 68.3                                 & 77.1                                 \\
\textit{img2img} & \ding{55}        & {\color[HTML]{FE0000} \textbf{54\%}}  & {\color[HTML]{333333} {\ul 62.4}}    & 71.7                                 \\
\textit{img2txt} & \textit{img2txt} & {\color[HTML]{036400} \textbf{34\%}}  & {\color[HTML]{000000} {\ul 67.1}}    & {\color[HTML]{000000} {\ul 76.2}}    \\
\rowcolor[HTML]{ECF4FF} 
\textit{img2txt} & \textit{img2img} & {\color[HTML]{036400} \textbf{34\%}}  & {\color[HTML]{340096} \textbf{68.3}} & {\color[HTML]{340096} \textbf{77.2}} \\ \bottomrule
\bottomrule[1pt]
\end{tabular}
\caption{\textbf{Ablation Study for the Importance Criteria}. \textbf{SHL-PM} and \textbf{DPL-PM} refer to the shallow-layer pruning module and deeper-layer pruning module, respectively. The {\color[HTML]{340096}\textbf{best}} model performance is highlighted in {\color[HTML]{340096}\textbf{blue}}, while degraded performance is underlined.}
\label{tab: ablation study}
\end{table}

\vspace{-0.15cm}
\section{Conclusion}
In this paper, we propose a hypothesis regarding visual information processing in MLLMs, suggesting that image tokens inject most visual information into instruction tokens in shallow layers while consolidate the remaining visual information in deeper layers. Results from information flow perturbation experiments confirm this hypothesis for the LLaVA-v1.5 series models. Building on these insights, we introduce Hierarchical Modality-Aware Pruning, a plug-and-play method that dynamically prunes image tokens at specific layers to improve inference speed. This method not only reaffirms our hypothesis but also demonstrates significant potential for practical applications.

\section*{Acknowledgements}
This work is supported by the National Natural Science Foundation of China under Grant 62176246. This work is
also supported by Anhui Province Key Research and Development Plan (202304a05020045),  Anhui Province Natural Science Foundation (2208085UD17) and National Natural Science Foundation of China under Grant 62406098.

{
    \small
    \bibliographystyle{ieeenat_fullname}
    \bibliography{main}
}

\clearpage
\setcounter{page}{1}
\maketitlesupplementary

\section{Related Work}
\textbf{Interpretability of LLMs.} Research on attention mechanisms has significantly enhanced our understanding of large language models. For instance, \citet{xiao2023efficient} highlight a phenomenon known as \textit{attention sink}, indicating that maintaining the key-value states of initial tokens can largely restore the performance of window attention, primarily due to the strong attention scores associated with these tokens. Furthermore, \citet{wang2023label} discovered that label words serve as anchors in in-context learning, facilitating the aggregation and distribution of task-relevant information. In addition, \citet{wu2024retrieval} identified a specific category of attention heads, referred to as retrieval heads, which are primarily responsible for extracting relevant information from lengthy contexts. However, most studies on attention mechanisms focus exclusively on text-based models, creating a gap in our understanding of information interaction within MLLMs. Our research aims to bridge this gap, offering new insights into how MLLMs process and utilize visual information.

\textbf{Inference Optimization for LLMs.} Research on efficient inference in large language models has primarily focused on two categories of optimization: (1) Memory Consumption Optimization, which includes methods such as FlashAttention \citep{dao2022flashattention}, vLLM \citep{kwon2023efficient}, and RingAttention \citep{liu2023ring} that enhance the memory efficiency of the attention module without significantly altering outcomes; and (2) Computation Simplification, which involves techniques like StreamingLLM and FastGen \citep{ge2023model} that improve inference efficiency by eliminating redundant attention calculations. This paper emphasizes the latter category. Most existing methods target text-only models, creating a notable gap in their applicability to MLLMs. Recent strategies, including FastV and VTW \citep{lin2024boosting}, have accelerated inference speeds through image token pruning, yet they overlook the shift in the dominant flow of visual information, failing to fully harness the potential for accelerating the inference of MLLMs.

\section{Results of modality impact assessment}
\label{appen: modality impact assessment}

\begin{figure*}[t]
    \centering
    \begin{subfigure}[t]{0.24\linewidth}
        \centering
        \includegraphics[width=\linewidth]{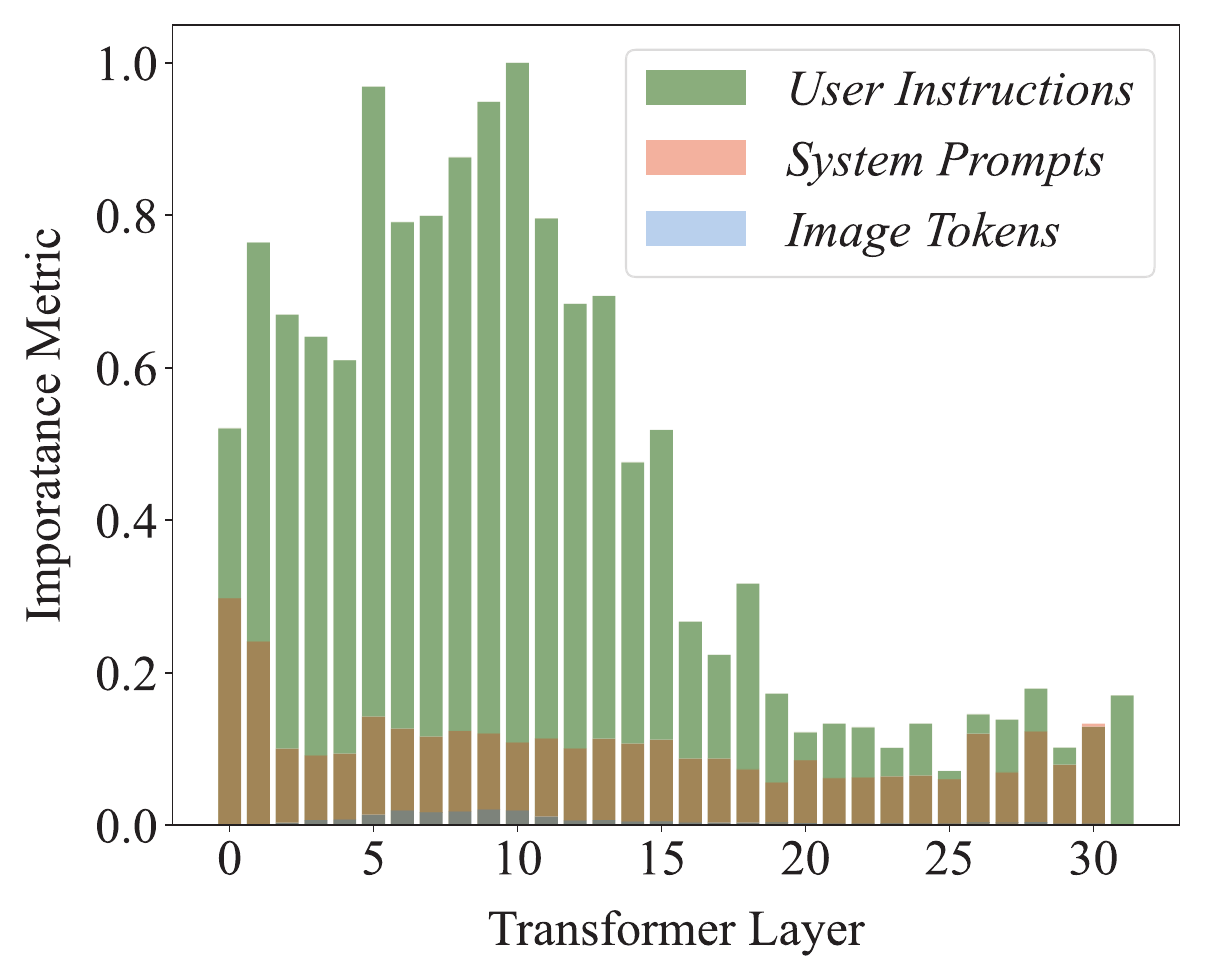}
        \caption{Sci-VQA (LLaVA-7B)}
        \label{fig: mod-ass scivqa-7b}
    \end{subfigure}
    \hfill
    \begin{subfigure}[t]{0.24\linewidth}
        \centering
        \includegraphics[width=\linewidth]{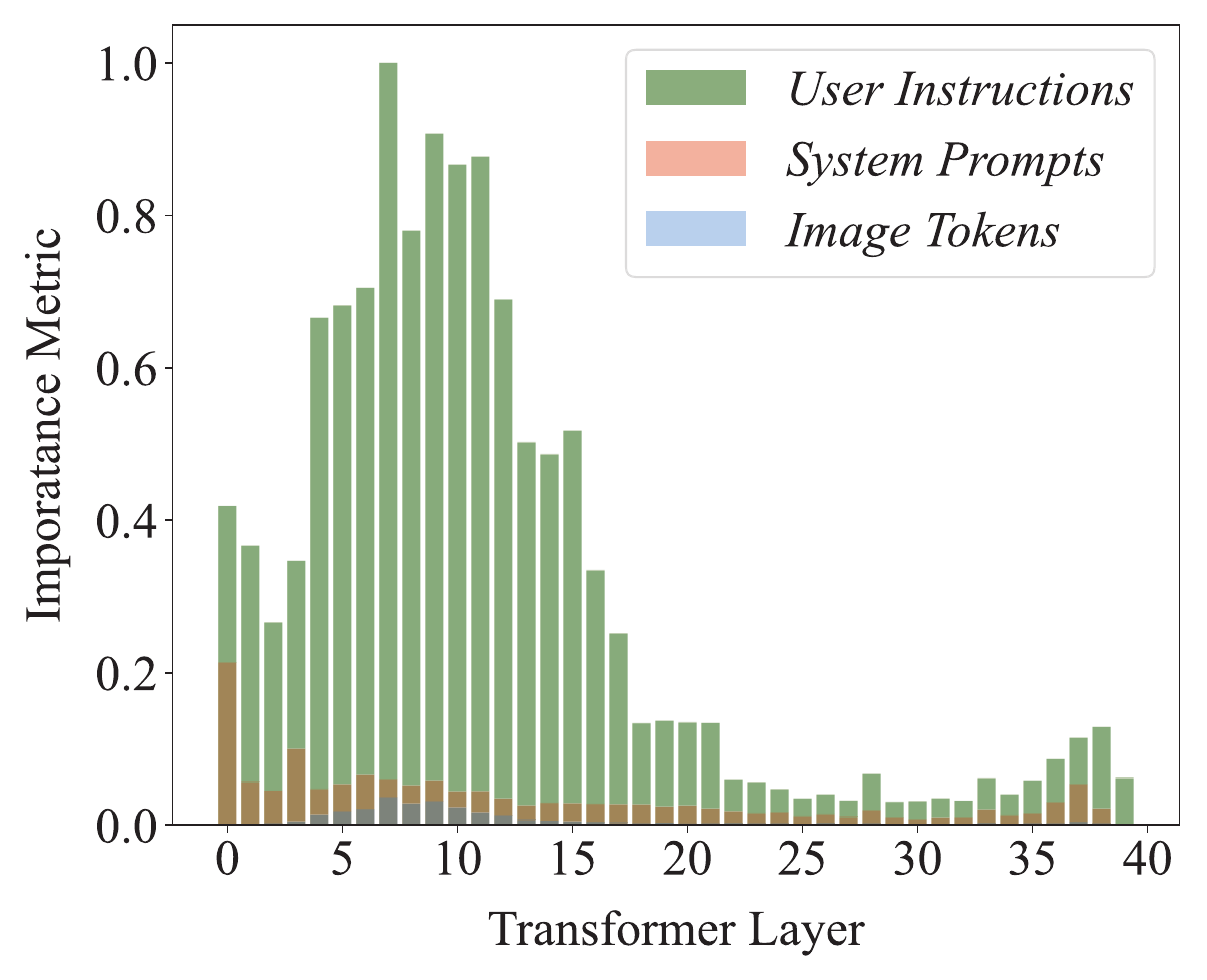}
        \caption{Sci-VQA (LLaVA-13B)}
        \label{fig: mod-ass scivqa-13b}
    \end{subfigure}
    \hfill
    \begin{subfigure}[t]{0.24\linewidth}
        \centering
        \includegraphics[width=\linewidth]{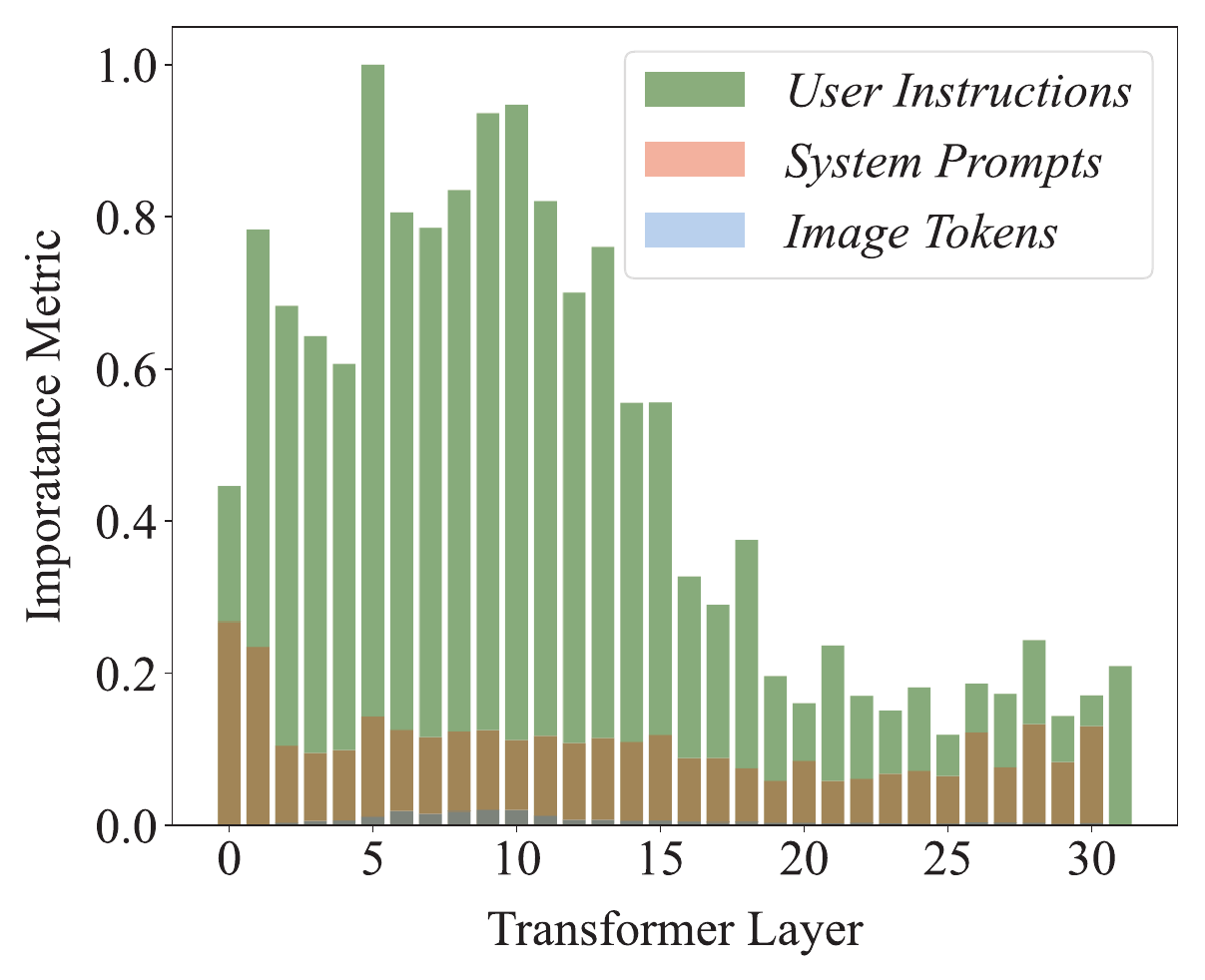}
        \caption{A-OKVQA (LLaVA-7B)}
        \label{fig: mod-ass aokvqa-7b}
    \end{subfigure}
    \hfill
    \begin{subfigure}[t]{0.24\linewidth}
        \centering
        \includegraphics[width=\linewidth]{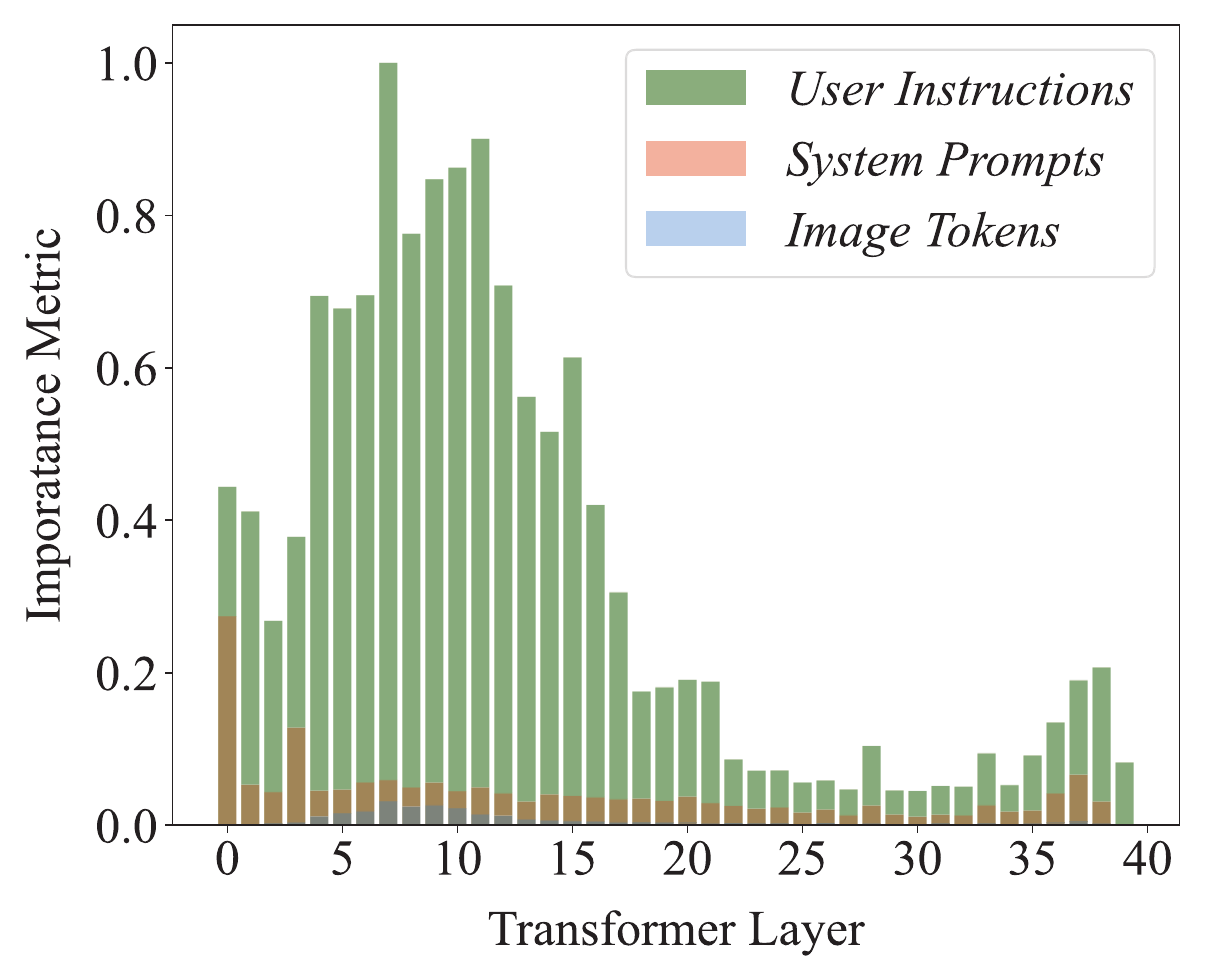}
        \caption{A-OKVQA (LLaVA-13B)}
        \label{fig: mod-ass aokvqa-13b}
    \end{subfigure}
    \caption{Experimental Results of modality impact assessment. The contribution of visual modality is lower than textual modality.}
    \label{fig: additional modality assessment}
\end{figure*}

\cref{fig: additional modality assessment} illustrates the influence of various modalities on the prediction outcomes of the LLaVA-1.5-7B and LLaVA-1.5-13B models within the Sci-VQA and AOKVQA datasets. 

\section{Results of visual flow analysis}
\label{appen: visual flow analysis}

\begin{figure*}[t]
    \centering
    \begin{subfigure}[t]{0.24\linewidth}
        \centering
        \includegraphics[width=\linewidth]{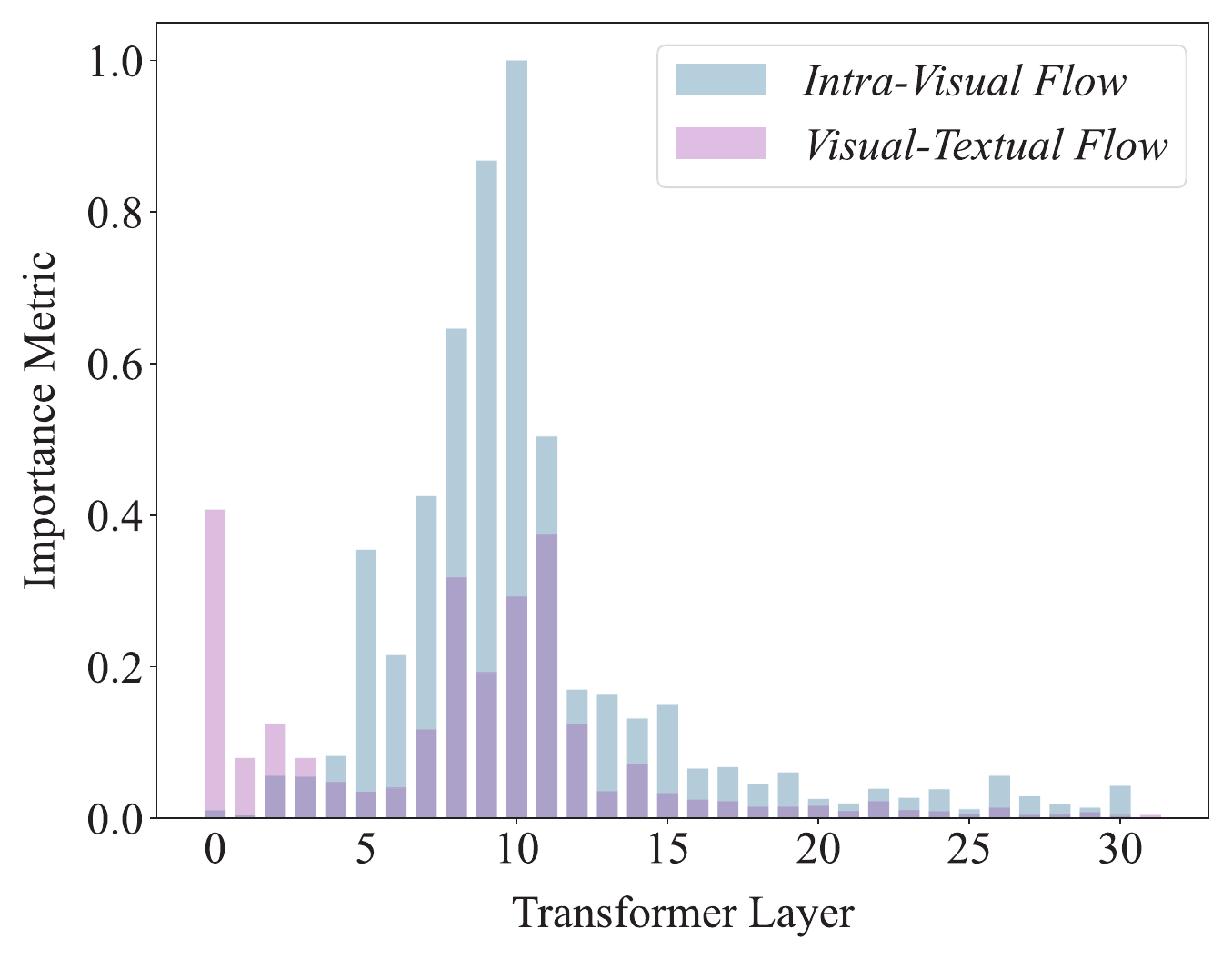}
        \caption{Sci-VQA (LLaVA-7B)}
        \label{fig: vis-ans scivqa-7b}
    \end{subfigure}
    \hfill
    \begin{subfigure}[t]{0.24\linewidth}
        \centering
        \includegraphics[width=\linewidth]{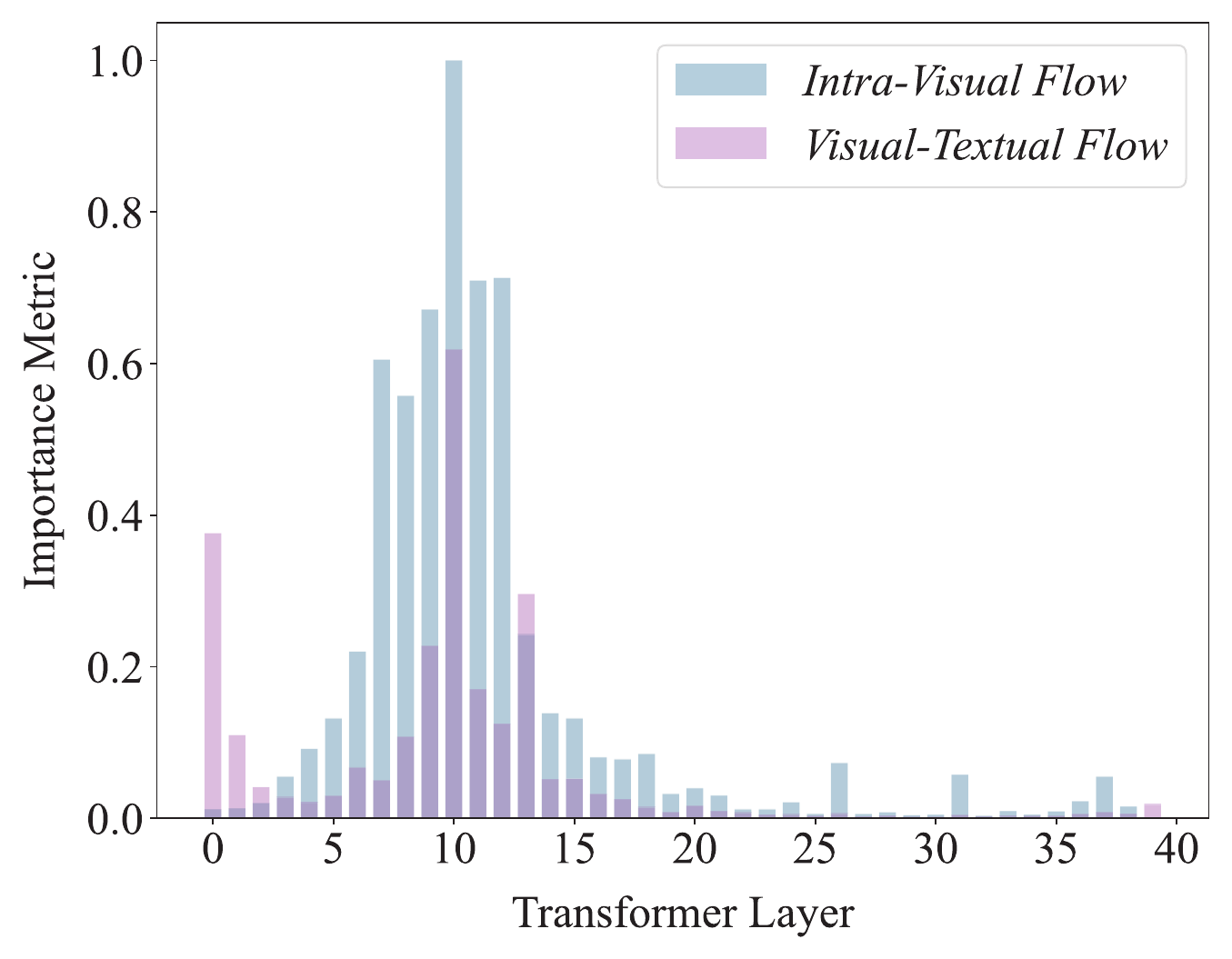}
        \caption{Sci-VQA (LLaVA-13B)}
        \label{fig: vis-ans scivqa-13b}
    \end{subfigure}
    \hfill
    \begin{subfigure}[t]{0.24\linewidth}
        \centering
        \includegraphics[width=\linewidth]{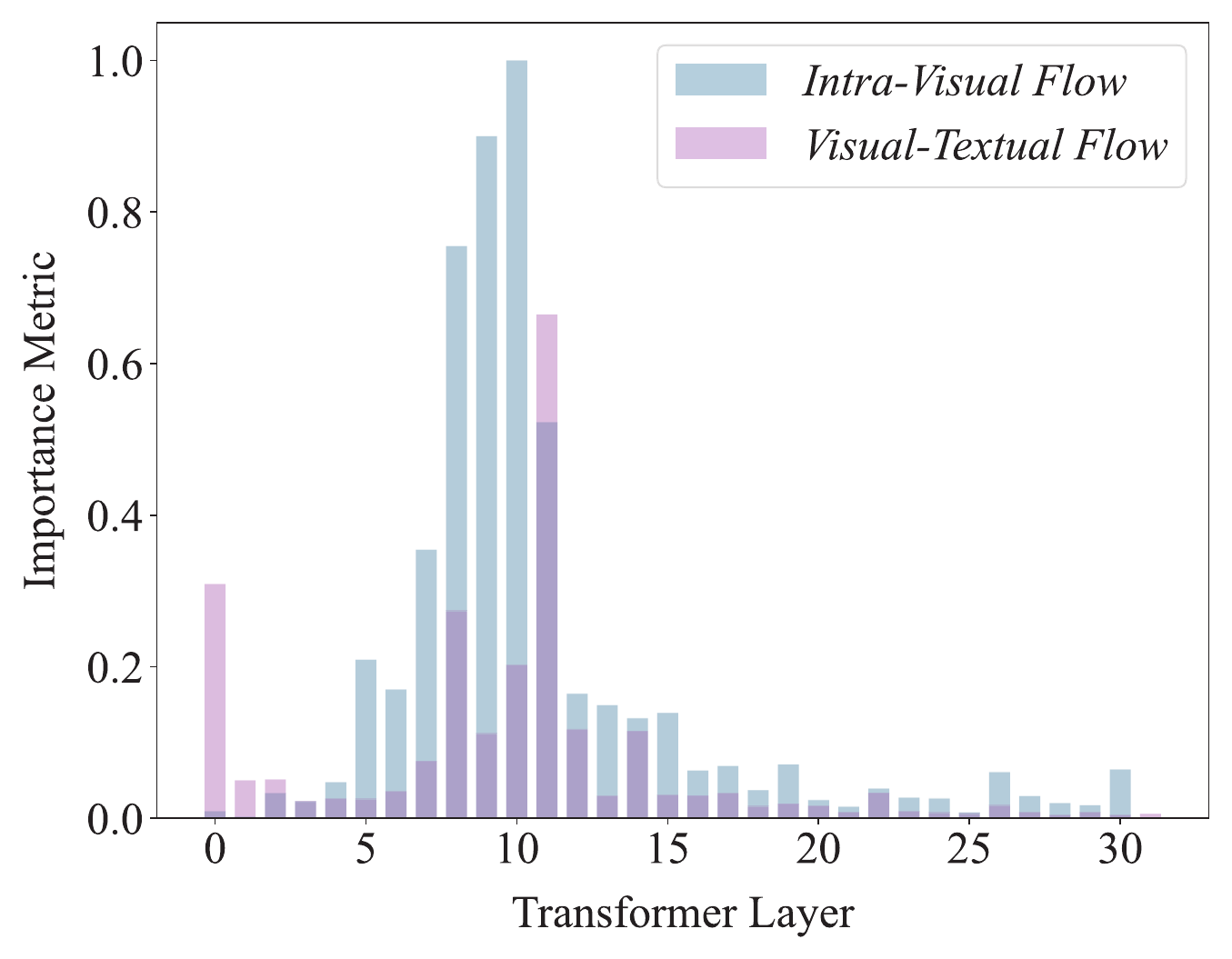}
        \caption{A-OKVQA (LLaVA-7B)}
        \label{fig: vis-ans aokvqa-7b}
    \end{subfigure}
    \hfill
    \begin{subfigure}[t]{0.24\linewidth}
        \centering
        \includegraphics[width=\linewidth]{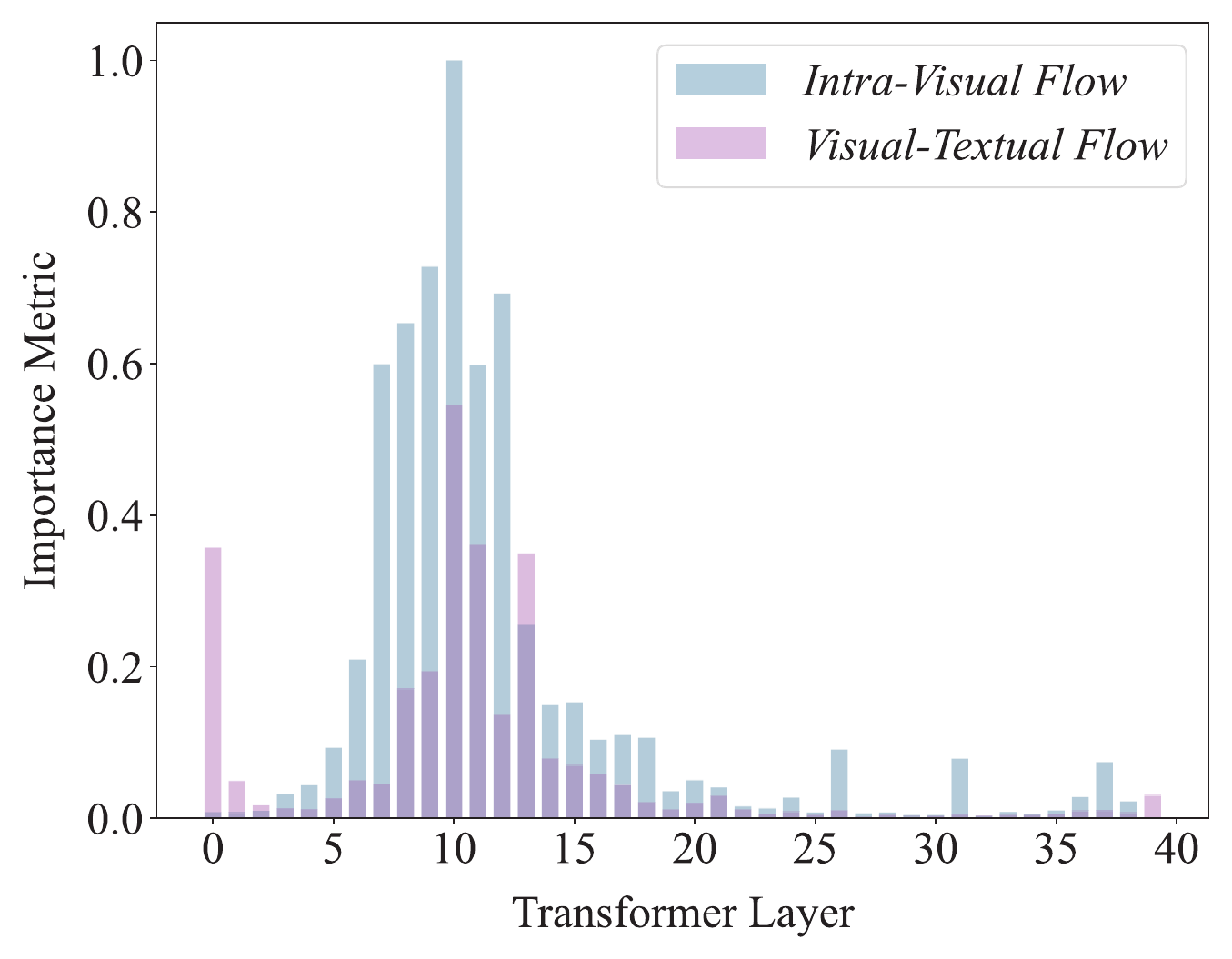}
        \caption{A-OKVQA (LLaVA-13B)}
        \label{fig: vis-ans aokvqa-13b}
    \end{subfigure}
    
    \caption{Additional Experimental Results of visual flow analysis. Dominant flow of visual information shifts as model depth increases.}
    \label{fig: additional visual flow analysis}
\end{figure*}

\cref{fig: additional visual flow analysis} illustrates the significance of intra-visual flow compared to visual-textual flow in the LLaVA-1.5-7B and LLaVA-1.5-13B models within the Sci-VQA and AOKVQA datasets. In shallow layers, the importance of \textit{visual-textual information flow} is notably high, while \textit{intra-visual information flow} is comparatively low. In deeper layers, \textit{intra-visual information flow} becomes dominant.

\section{Details for computation of prediction bias}
\label{appen: computation of E_vv & E_vt}

The prediction biases, $E_{vv,l}$ and $E_{vt,l}$, resulting from disruptions in \textit{visual-textual} and \textit{intra-visual information flows}, as introduced in \cref{subsec: deeper layers}, may have caused confusion. Here, we provide a more detailed explanation of their calculation methods.

In the absence of disruption to information flow, \textbf{Score Consistency} of the model is denoted as $C$. When \textit{intra-visual information flow} in the $l$-th layer is sidrupted, \textbf{Score Consistency} is represented as $C_{vv,l}$. The prediction bias $E_{vv,l}$ resulting from this disruption is calculated as follows: 
\begin{equation}
    E_{vv,l} = C - C_{vv,l}.
\end{equation}
Similarly, \textbf{Score Consistency} of the model after disrupting \textit{visual-textual information flow} in the $l$-th layer is denoted as $C_{vt,l}$. Consequently, the prediction bias $E_{vt,l}$ resulting from this disruption is calculated as follows:
\begin{equation}
    E_{vt,l} = C - C_{vt,l}.
\end{equation}

\section{Reasons for Using Bias Ratio}
\label{appen: evaluation metrics D_l}

We use the $D_l$ metric to validate the importance of the intra-visual information flow, based on two main considerations: 
\begin{itemize}[left=0.3cm, itemsep=0.1pt, topsep=0.1pt]
    \item As demonstrated by the experimental results in \cref{subsec: visual modality impact assessment}, the prediction outcomes are primarily influenced by intra-textual information flow, which weakens as the network depth increases. Consequently, although intra-visual information flow becomes more prominent in deeper layers, its disruption has minimal impact on prediction outcomes. Therefore, we use the significance of visual-textual information flow as a baseline and apply a logarithmic ratio to measure the variation in the importance of intra-visual information flow.
    \item We focus on the relative strength between intra-visual and visual-textual information flows to clearly illustrate the shift in the mechanism of visual information processing in Multimodal large language models.
\end{itemize}

\section{Experimental Results for HiMAP}
\label{appen: extended experimental results}

\Cref{appen: ChartQA Doc-QA} discusses the performance on ChartQA and DocQA datasets, \Cref{appen: mme} presents results on the MME Benchmark, \Cref{appen: case study} analyzes HiMAP's impact on content generation in LLaVA-Bench, and \Cref{appen: inference speed} compares the inference speed improvements between HiMAP and FastV.

\begin{table}[h]
\centering
\begin{tabular}{@{}c|c|c|cc@{}}
\toprule[1pt]
\toprule
\textbf{Model}                                    & \textbf{Method} & \textbf{Ratio}                                               & \textbf{ChartQA}                                             & \textbf{DocQA}                                               \\ \midrule
{\color[HTML]{333333} }                           & Baseline        & 100\%                                                        & {\color[HTML]{CB0000} \textbf{9.7}}                          & 8.6                                                          \\
\multirow{-2}{*}{{\color[HTML]{333333} LLaVA-7B}} & HiMAP           & \cellcolor[HTML]{EFEFEF}{\color[HTML]{036400} \textbf{24\%}} & \cellcolor[HTML]{EFEFEF}9.4                                  & \cellcolor[HTML]{EFEFEF}{\color[HTML]{CB0000} \textbf{8.8}}  \\ \midrule
                                                  & Baseline        & 100\%                                                        & 65                                                           & 64.9                                                         \\
\multirow{-2}{*}{QwenVL-7B}                         & HiMAP           & \cellcolor[HTML]{EFEFEF}{\color[HTML]{036400} \textbf{23\%}} & \cellcolor[HTML]{EFEFEF}{\color[HTML]{CB0000} \textbf{65.1}} & \cellcolor[HTML]{EFEFEF}{\color[HTML]{CB0000} \textbf{65.3}} \\ \bottomrule
\bottomrule[1pt]
\end{tabular}
\caption{\textbf{Performance on ChartQA and DocQA datasets.} In each configuration, the {\color[HTML]{CB0000} \textbf{highest scores}} are highlighted in {\color[HTML]{CB0000} \textbf{red}}, while the {\color[HTML]{036400} \textbf{lowest computational cost}} are marked in {\color[HTML]{036400} \textbf{green}}. The parameters for HiMAP are set as $K_1=2$, $R_1=50\%$, $K_2=8$, and $R_2=75\%$.}
\label{tab: chartqa docqa}
\end{table}

\subsection{Results on ChartQA \& Doc-QA}
\label{appen: ChartQA Doc-QA}

We conducted a comprehensive evaluation of HiMAP's performance on the ChartQA \cite{masry2022chartqabenchmarkquestionanswering} and DocQA \cite{mathew2021docvqadatasetvqadocument} datasets, utilizing the LLaVA-v1.5 model family as our foundation. The experimental results, summarized in \cref{tab: chartqa docqa}, demonstrate that effectively reduces computational overhead with minimal loss in model performance. This highlights HiMAP's efficacy in fine-grained visual question-answering tasks, showing that its pruning of visual tokens does not compromise the model's ability to perceive image details.

\subsection{Results on MME}
\label{appen: mme}

\cref{tab: mme} illustrates the experimental outcomes of LLaVA-v1.5-7B model on the MME benchmark after incorporating HiMAP. The results indicate that, for both perception-focused and cognition-focused tasks, HiMAP method not only significantly reduces computational costs but also preserves or marginally enhances the model’s performance.

\begin{table}[h]
\centering
\begin{tabular}{@{}c|c|c|cc@{}}
\toprule[1pt]
\toprule
\textbf{Model}              & \textbf{Method} & \textbf{Ratio}                                               & \textbf{MME-P}                                        & \textbf{MME-C}                                                \\ \midrule
                            & Baseline        & 100\%                                                        & 1459.2                                                         & 290.7                                                         \\
\multirow{-2}{*}{LLaVA-7B}  & HiMAP           & \cellcolor[HTML]{EFEFEF}{\color[HTML]{036400} \textbf{24\%}} & \cellcolor[HTML]{EFEFEF}{\color[HTML]{CB0000} \textbf{1492.6}} & \cellcolor[HTML]{EFEFEF}{\color[HTML]{CB0000} \textbf{292.5}} \\ \midrule
                            & Baseline        & 100\%                                                        & 1517.3                                                         & 277.1                                                         \\
\multirow{-2}{*}{LLaVA-13B} & HiMAP           & \cellcolor[HTML]{EFEFEF}{\color[HTML]{036400} \textbf{23\%}} & \cellcolor[HTML]{EFEFEF}{\color[HTML]{CB0000} \textbf{1526.9}} & \cellcolor[HTML]{EFEFEF}{\color[HTML]{CB0000} \textbf{282.5}} \\ \bottomrule
\bottomrule[1pt]
\end{tabular}
\vspace{0.25cm}
\caption{\textbf{Performance on MME Benchmark.} In each configuration, the {\color[HTML]{CB0000} \textbf{highest scores}} are highlighted in {\color[HTML]{CB0000} \textbf{red}}, while the {\color[HTML]{036400} \textbf{lowest computational cost}} are marked in {\color[HTML]{036400} \textbf{green}}. The parameters for HiMAP are set as $K_1=2$, $R_1=50\%$, $K_2=8$, and $R_2=75\%$.}
\label{tab: mme}
\end{table}

\begin{figure*}[t]
    \centering
    \begin{subfigure}[b]{\textwidth}
        \centering
        \includegraphics[width=0.75\textwidth]{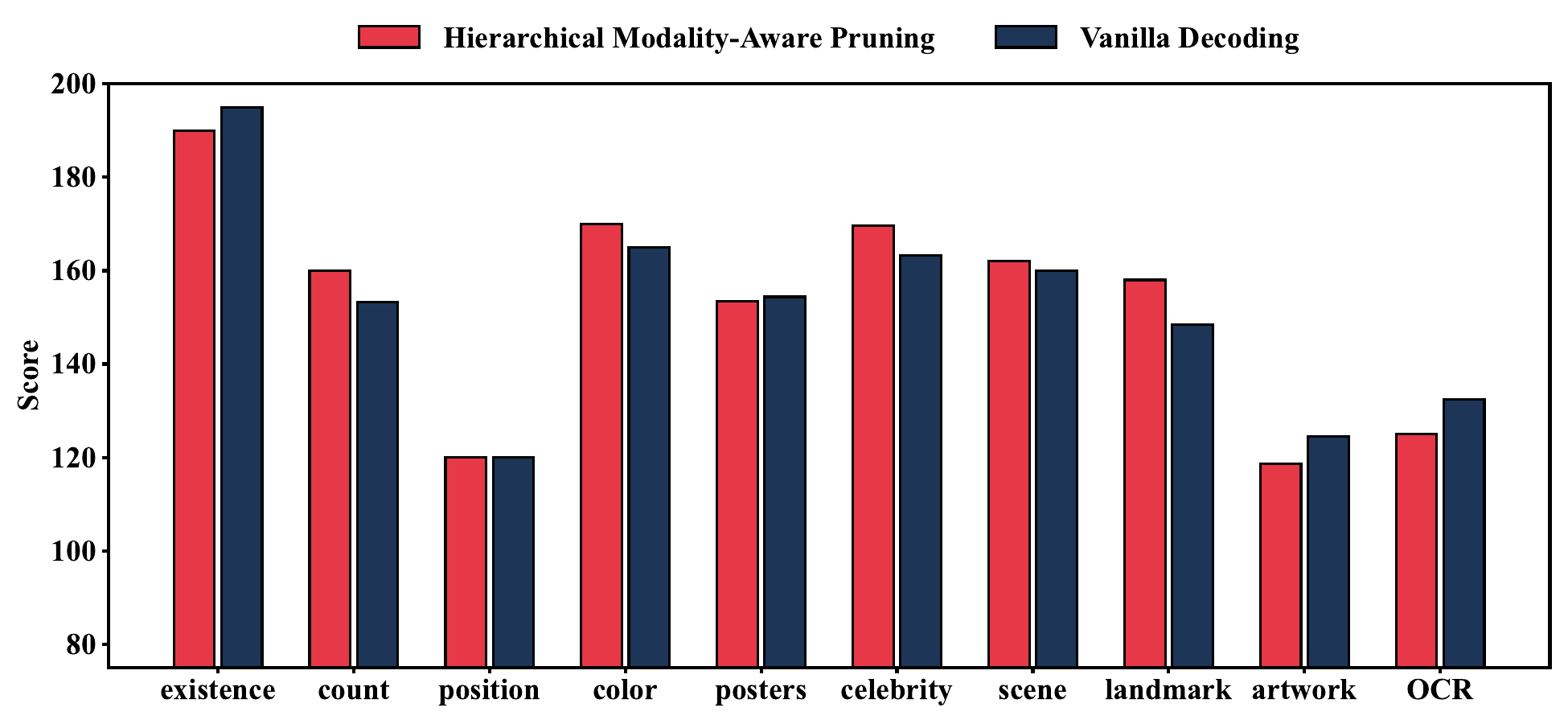}
        \caption{Perception-Related Tasks from MME Benchmark}
        \label{fig: mmep}
    \end{subfigure}
    \vspace{0.2cm} 
    \begin{subfigure}[b]{\textwidth}
        \centering
        \includegraphics[width=0.75\textwidth]{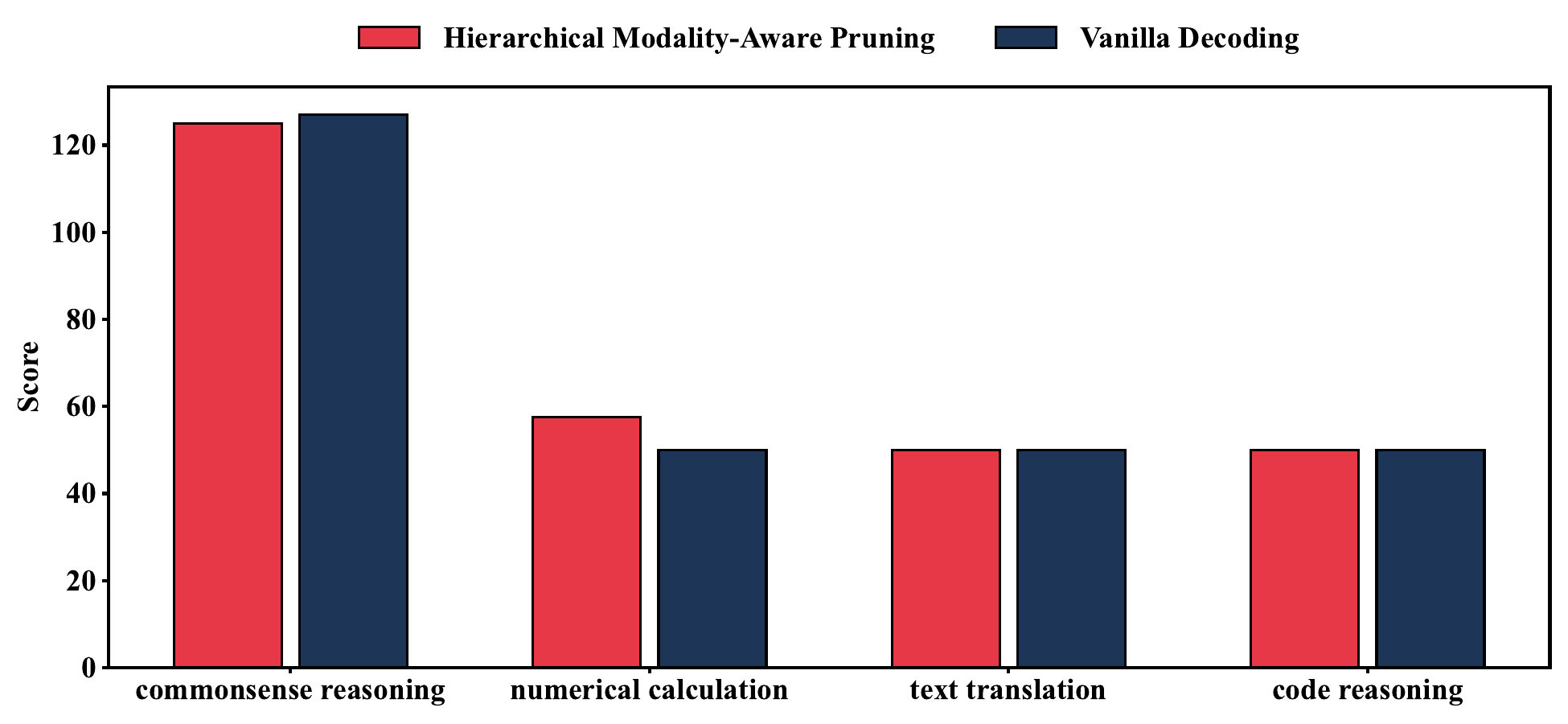}
        \caption{Cognition-Related Tasks from MME Benchmark}
        \label{fig: mmec}
    \end{subfigure}
    \vspace{-0.75cm}
    \caption{\textbf{Performance of LLaVA-v1.5-13B model on the MME Benchmark.} After applying HiMAP method, the model retained nearly all of its original performance across each task.}
    \label{fig: mme}
\end{figure*}

 \cref{fig: mme} highlights the performance of the LLaVA-v1.5-7B model on each subtask of the MME Benchmark after applying HiMAP, demonstrating that HiMAP effectively sustains the model’s performance across all subtasks.

\begin{table*}[t]
\centering
\begin{tabular}{@{}c|c|cccc@{}}
\toprule[1pt]
\toprule
\textbf{Model}                   & \textbf{Method} & \textbf{Accurancy}           & \textbf{Total Time}                                          & \textbf{GPU-Memory}                                         & \textbf{Latency/Example}                                       \\ \midrule
                                 & Baseline        & 67.9                         & 6:36                                                         & 17G                                                         & 0.197s                                                         \\
                                 & FastV           & 67.8                         & 4:51                                                         & 15G                                                         & 0.144s                                                         \\
\multirow{-3}{*}{LLaVA-v1.5-7B}  & HiMAP           & \cellcolor[HTML]{EFEFEF}68.2 & \cellcolor[HTML]{EFEFEF}{\color[HTML]{036400} \textbf{3:54}} & \cellcolor[HTML]{EFEFEF}{\color[HTML]{036400} \textbf{14G}} & \cellcolor[HTML]{EFEFEF}{\color[HTML]{036400} \textbf{0.116s}} \\ \midrule
                                 & Baseline        & 71.6                         & 10:45                                                        & 31G                                                         & 0.320s                                                         \\
                                 & FastV           & 71.2                         & 7:13                                                         & 26G                                                         & 0.214s                                                         \\
\multirow{-3}{*}{LLaVA-v1.5-13B} & HiMAP           & \cellcolor[HTML]{EFEFEF}72.5 & \cellcolor[HTML]{EFEFEF}{\color[HTML]{036400} \textbf{5:13}} & \cellcolor[HTML]{EFEFEF}{\color[HTML]{036400} \textbf{23G}} & \cellcolor[HTML]{EFEFEF}{\color[HTML]{036400} \textbf{0.158s}} \\ \bottomrule
\bottomrule[1pt]
\end{tabular}
\vspace{0.25cm}
\caption{\textbf{Comparison of inference speed and GPU memory usage between HiMAP and FastV.} HiMAP outperforms FastV by delivering faster inference speeds and lower GPU memory usage while maintaining higher prediction accuracy. In each configuration, the {\color[HTML]{036400} \textbf{fastest inference speed}} and the {\color[HTML]{036400} \textbf{lowest GPU memory usage}} are highlighted in {\color[HTML]{036400} \textbf{green}}.}
\label{tab: inference speed}
\end{table*}

\subsection{Case Study on LLaVA-Bench}
\label{appen: case study}

\begin{figure*}[t]  
    \centering     
    \includegraphics[width=0.85\linewidth]{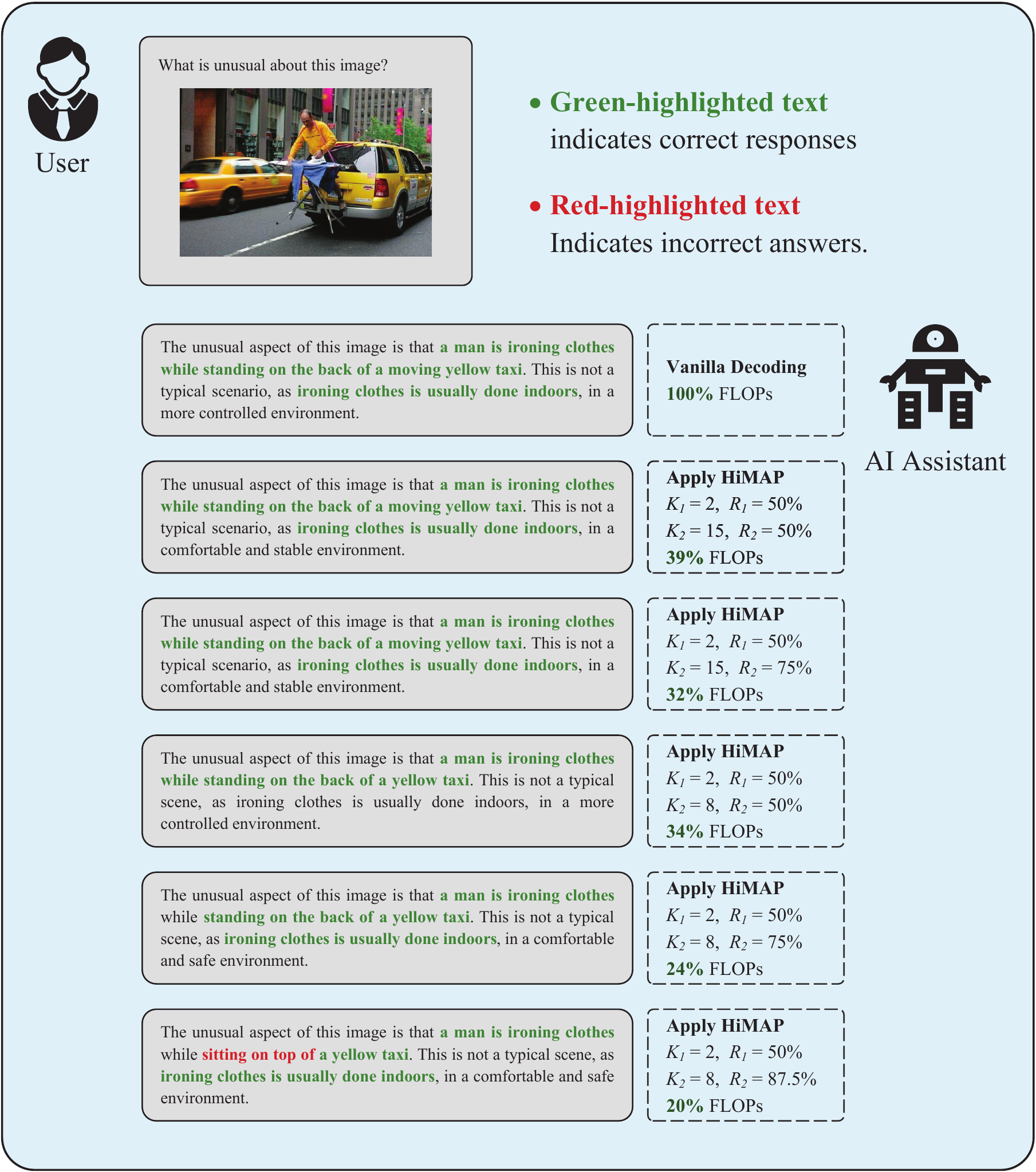} 
    \caption{The output results after applying HiMAP method. {\color[HTML]{036400}\textbf{Correct segments}} of outputs are highlighted in {\color[HTML]{036400}\textbf{green}}, while {\color[HTML]{CB0000}\textbf{incorrect segments}} are marked in {\color[HTML]{CB0000}\textbf{red}}. The findings indicate that HiMAP does not compromise the quality of the responses generated by the model.} 
    \label{fig: outputs after applying HiMAP} 
\end{figure*}

\cref{fig: outputs after applying HiMAP} illustrates the long-text generation performance of the LLaVA-v1.5-7B model on LLaVA-Bench after the application of HiMAP. The results indicate that, when HiMAP is configured with appropriate parameters, it achieves a substantial reduction in computational overhead while almost entirely retaining the model’s ability to produce high-quality long-text responses. Remarkably, even with aggressive parameter settings ($K_1=2$, $R_1=50\%$, $K_2=8$, $R_2=87.5\%$), the model consistently delivers fluent and accurate outputs.

\subsection{Comparison of Inference Speeds}
\label{appen: inference speed}

We utilized the LLaVA-v1.5 model family to evaluate the inference speed and GPU memory usage of HiMAP and FastV on the ScienceQA dataset. The results, presented in \cref{tab: inference speed}, show that applying HiMAP achieves higher prediction accuracy, faster inference speed, and lower GPU memory consumption compared to FastV. These improvements are primarily driven by HiMAP’s ability to perform precise and efficient pruning of visual tokens. By leveraging different vision-dominant information streams at the model's shallow and deep layers, HiMAP maximizes the potential for inference acceleration.

\begin{figure*}[t]
    \centering
    \begin{subfigure}[b]{0.43\textwidth}
        \centering
        \includegraphics[width=\textwidth]{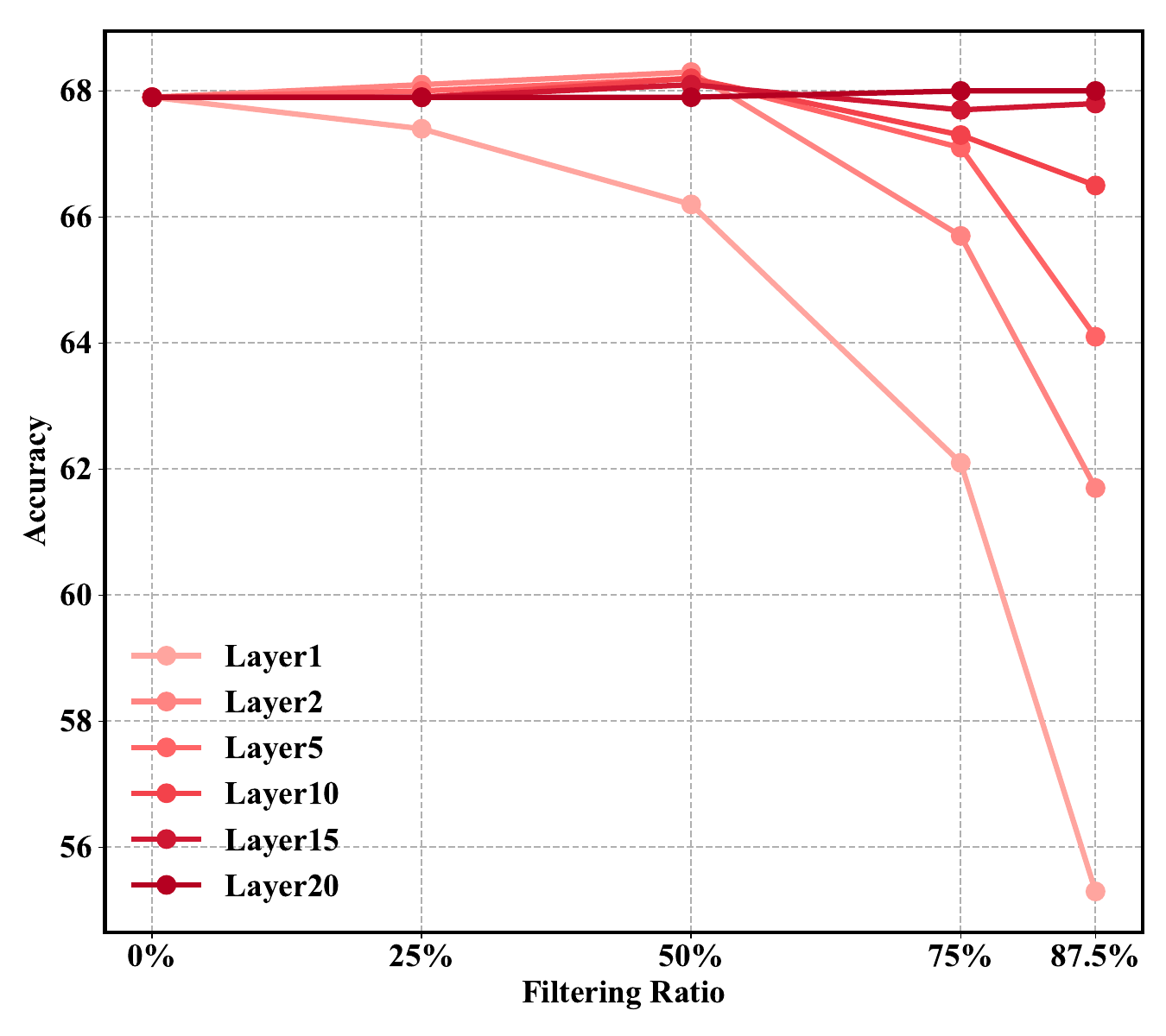}
        \caption{Accuracy}
        \label{fig: k1r1 accurancy}
    \end{subfigure}
    \hspace{0.05\textwidth}
    \begin{subfigure}[b]{0.43\textwidth}
        \centering
        \includegraphics[width=\textwidth]{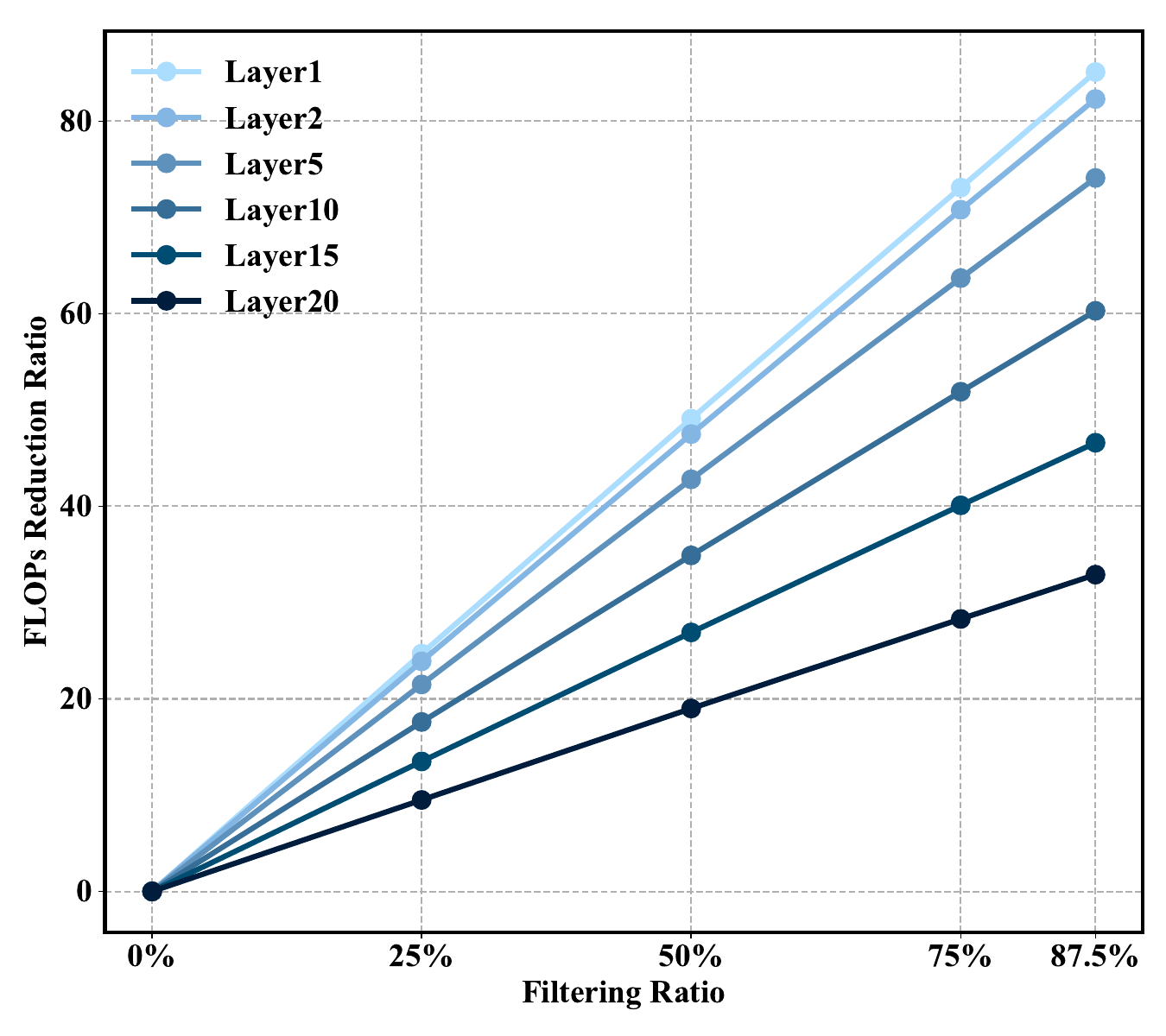}
        \caption{FLOPs Reduction Ratio}
        \label{fig: k1r1 reduction}
    \end{subfigure}
    \caption{\textbf{Ablation Study on the Parameters $K_1$ and $R_1$.}}
    \label{fig: k1r1}
\end{figure*}

\begin{table*}[t]
\small
\centering
\begin{tabular}{@{}c|cc|cc|cccc@{}}
\toprule[1pt]
\toprule
\textbf{Model}                   & $\boldsymbol{K_2}$ & $\boldsymbol{R_2}$ & \textbf{TFLOPs} & \textbf{FLOPs Ratio} & \textbf{ScienceQA}                   & \textbf{A-OKVQA}                     & \textbf{NoCaps}                      & \textbf{Flickr30k}                   \\ \midrule
                                 & \multicolumn{2}{c|}{Baseline} & 2.98            & 100\%                & 67.9                                 & 76.7                                 & 78.8                                 & 50.9                                 \\
                                 & 8             & 87.5\%        & 0.59            & 20\%                 & 68                                   & 71.7                                 & 74.6                                 & 46.2                                 \\
                                 & 8             & 75\%          & 0.73            & 24\%                 & {\color[HTML]{CB0000} \textbf{68.3}} & {\color[HTML]{CB0000} \textbf{77.2}} & 76.1                                 & 47.2                                 \\
                                 & 8             & 50\%          & 1.01            & 33\%                 & {\color[HTML]{333333} 67.8}          & {\color[HTML]{00009B} \textbf{76.7}} & 76.9                                 & 49.1                                 \\
                                 & 15            & 87.5\%        & 0.88            & 29\%                 & 68.1                                 & 77.2                                 & 77.5                                 & 50.1                                 \\
                                 & 15            & 75\%          & 0.97            & 32\%                 & 68.2                                 & 77.2                                 & {\color[HTML]{00009B} \textbf{78.7}} & {\color[HTML]{00009B} \textbf{51.3}} \\
\multirow{-7}{*}{LLaVA-v1.5-7B}  & 15            & 50\%          & 1.17            & 39\%                 & {\color[HTML]{00009B} \textbf{68.2}} & 77.2                                 & {\color[HTML]{CB0000} \textbf{79.2}} & {\color[HTML]{CB0000} \textbf{51.7}} \\ \midrule
                                 & \multicolumn{2}{c|}{Baseline} & 5.81            & 100\%                & 71.6                                 & 82                                   & 82.8                                 & 53.6                                 \\
                                 & 8             & 87.5\%        & 1.08            & 18\%                 & 72                                   & 79.9                                 & 77.5                                 & 47.5                                 \\
                                 & 8             & 75\%          & 1.36            & 23\%                 & {\color[HTML]{00009B} \textbf{72.1}} & {\color[HTML]{CB0000} \textbf{81.4}} & 82.5                                 & 52.6                                 \\
                                 & 8             & 50\%          & 1.94            & 33\%                 & 71.9                                 & {\color[HTML]{00009B} \textbf{81.2}} & 82.7                                 & 52.8                                 \\
                                 & 15            & 87.5\%        & 1.52            & 26\%                 & 71.7                                 & 81                                   & 82.9                                 & 52.6                                 \\
                                 & 15            & 75\%          & 1.74            & 30\%                 & {\color[HTML]{CB0000} \textbf{72.5}} & 81.2                                 & {\color[HTML]{303498} \textbf{83.7}} & {\color[HTML]{303498} \textbf{53.8}} \\
\multirow{-7}{*}{LLaVA-v1.5-13B} & 15            & 50\%          & 2.19            & 37\%                 & 72.1                                 & 81.1                                 & {\color[HTML]{CB0000} \textbf{83.9}} & {\color[HTML]{CB0000} \textbf{54.1}} \\ \bottomrule
\bottomrule[1pt]
\end{tabular}
\caption{\textbf{Ablation Study on $K_2$ and $R_2$.} In each configuration, the {\color[HTML]{CB0000} \textbf{highest score}} is marked in {\color[HTML]{CB0000} \textbf{red}}, while the {\color[HTML]{00009B} \textbf{second-highest score}} is marked in {\color[HTML]{00009B} \textbf{blue}}.}
\label{tab: k2r2}
\end{table*}

\section{Ablation Studies on HiMAP}
\label{appen: ablation study}

\cref{appen: k & R} delves into the impact of HiMAP's parameters, $K$ and $R$, on pruning performance. \cref{appen: pruning module} evaluates the individual contributions of the shallow-layer and deeper-layer pruning modules to predictions.

\subsection{Effect of Filtering Layer \& Filtering Ratio}
\label{appen: k & R}

Ablation studies were performed on parameters $K_1$ and $R_1$. After excluding the deeper-layer pruning module, we tuned $K_1$ and $R_1$ to assess their influence on HiMAP's pruning effectiveness. As illustrated in \cref{fig: k1r1}, it is clear that pruning less than $50 \%$ of visual tokens beyond the second model layer does not substantially impact prediction accuracy.

We conducted further ablation experiments on the parameters $K_2$ and $R_2$. By fixing $K_1=2$ and $R_1=50\%$, we adjusted the values of $K_2$ and $R_2$ to analyze their impact on the performance of HiMAP pruning. The experimental results are presented in Table 1. For short-text response generation tasks, such as ScienceQA and A-OKVQA, setting $K_2=8$ and $R_2=75\%$ effectively minimizes computational overhead while maintaining model performance. However, for long-text response generation tasks, such as Nocaps and Flickr30k, a more conservative configuration, $K_2=15$ and $R_2=75\%$, is necessary to ensure the model's performance remains uncompromised.

\subsection{Effect of Pruning Module}
\label{appen: pruning module}

\cref{tab: shallow deeper} presents the results of ablation studies conducted on the pruning modules. It is evident that applying either the shallow-layer or deeper-layer pruning module individually can reduce computational overhead without compromising model performance. This demonstrates that both modules effectively accelerate model inference.

\begin{table*}[t]
\small
\centering
\begin{tabular}{@{}c|cc|cc|cc@{}}
\toprule[1pt]
\toprule
\textbf{Model}                  & \textbf{SHL-PM}           & \textbf{DPL-PM}            & \textbf{TFLOPs} & \textbf{FLOPs Ratio} & \textbf{ScienceQA} & \textbf{A-OKVQA} \\ \midrule
\multirow{4}{*}{LLaVA-v1.5-7B}  & \ding{55} & \ding{55} & 2.98            & 100\%                & 67.9               & 76.6             \\
                                & \ding{51} & \ding{55} & 1.56            & 54\%                 & \textbf{68.3}      & \textbf{77.1}    \\
                                & \ding{55} & \ding{51} & 1.78            & 34\%                 & \textbf{68.1}      & \textbf{77.2}    \\
                                & \ding{51} & \ding{51} & 0.73            & 24\%                 & 68.3               & 77.2             \\ \midrule
\multirow{4}{*}{LLaVA-v.15-13B} & \ding{55} & \ding{55} & 5.81            & 100\%                & 71.6               & 82.0             \\
                                & \ding{51} & \ding{55} & 3.09            & 53\%                 & \textbf{71.8}      & 81.2             \\
                                & \ding{55} & \ding{51} & 1.73            & 30\%                 & \textbf{72.0}      & 81.3             \\
                                & \ding{51} & \ding{51} & 1.36            & 23\%                 & 72.1               & 81.4             \\ \bottomrule
\bottomrule[1pt]
\end{tabular}
\caption{\textbf{Ablation Study on Shallow-layer Pruning Module and Deeper-layer Pruning Module.}}
\vspace{0.25cm}
\label{tab: shallow deeper}
\end{table*}
\vspace{-0.2cm}

\section{Prompts for different tasks}
\label{appen: prompts for different tasks}

\textbf{Sci-VQA Dataset.} In the Sci-VQA dataset, input template for the model is presented below, with the prompts highlighted in {\color[HTML]{036400}\textbf{green}} and the image highlighted in {\color[HTML]{CB0000}\textbf{red}}.

\vspace{0.15cm}
\begin{tcolorbox}[title=\textbf{Sci-VQA Dataset}, colback=gray!10, colframe=black, boxrule=0.5mm]

A chat between a curious user and an artificial intelligence assistant. The assistant gives helpful, detailed, and polite answers to the user's questions.

\vspace{0.2cm}
\begin{hangparas}{1.22cm}{1}
\textbf{USER: } {\color[HTML]{CB0000}\textbf{IMAGE}} \\
{\color[HTML]{036400}\textbf{Context: Select the best answer.}} \\
Which property do these three objects have in common? \\
A. shiny B. slippery C. opaque \\
{\color[HTML]{036400}\textbf{Answer with the option's letter from the given choices directly.}}
\end{hangparas}

\vspace{0.2cm}
\textbf{ASSISTANT:}

\end{tcolorbox}
\vspace{0.15cm}

\noindent \textbf{AOKVQA Dataset.} In the AOKVQA dataset, input template for the model is presented below, with the prompts highlighted in {\color[HTML]{036400}\textbf{green}} and the image highlighted in {\color[HTML]{CB0000}\textbf{red}}. 

\vspace{0.15cm}
\begin{tcolorbox}[title=\textbf{A-OKVQA Dataset}, colback=gray!10, colframe=black, boxrule=0.5mm]

A chat between a curious user and an artificial intelligence assistant. The assistant gives helpful, detailed, and polite answers to the user's questions.

\vspace{0.2cm}
\begin{hangparas}{1.22cm}{1}
\textbf{USER: } {\color[HTML]{CB0000}\textbf{IMAGE}} \\ 
{\color[HTML]{036400}\textbf{Analyse the image and choose the best answer for the following question:}} \\ 
What is in the motorcyclist's mouth? \\ 
Options: (A) toothpick (B) food (C) popsicle stick (D) cigarette \\ 
{\color[HTML]{036400}\textbf{Output the letter of the correct answer.}}
\end{hangparas}

\vspace{0.2cm}
\textbf{ASSISTANT:}
\end{tcolorbox}
\vspace{0.15cm}

\noindent \textbf{POPE Benchmark.} In the POPE benchmark, input template for the model is presented below, with the prompts highlighted in {\color[HTML]{036400}\textbf{green}} and the image highlighted in {\color[HTML]{CB0000}\textbf{red}}.

\vspace{0.15cm}
\begin{tcolorbox}[title=\textbf{POPE Benchmark}, colback=gray!10, colframe=black, boxrule=0.5mm]

A chat between a curious user and an artificial intelligence assistant. The assistant gives helpful, detailed, and polite answers to the user's questions.

\vspace{0.2cm}
\begin{hangparas}{1.22cm}{1}
\textbf{USER: } {\color[HTML]{CB0000}\textbf{IMAGE}} \\
{\color[HTML]{036400}\textbf{Is there a cow in the image? Please just answer yes or no.}}
\end{hangparas}

\vspace{0.2cm}
\textbf{ASSISTANT:}
\end{tcolorbox}
\vspace{0.15cm}

\noindent \textbf{Nocaps \& Flickr30k Datasets.} In the Nocaps and Flickr30k dataset, input template for the model is presented below, with the prompts highlighted in {\color[HTML]{036400}\textbf{green}} and the image highlighted in {\color[HTML]{CB0000}\textbf{red}}.

\vspace{0.15cm}
\begin{tcolorbox}[title=\textbf{Nocaps \& Flickr30k Datasets}, colback=gray!10, colframe=black, boxrule=0.5mm]

A chat between a curious user and an artificial intelligence assistant. The assistant gives helpful, detailed, and polite answers to the user's questions.

\vspace{0.2cm}
\begin{hangparas}{1.22cm}{1}
\textbf{USER: } {\color[HTML]{CB0000}\textbf{IMAGE}} \\
{\color[HTML]{036400}\textbf{Provide a one-sentence caption for the provided image.}}
\end{hangparas}

\vspace{0.2cm}
\textbf{ASSISTANT:}
\end{tcolorbox}
\vspace{0.15cm}

\end{document}